%% file: paper_arxiv.tex
\documentclass{article}

\usepackage{arxiv}
\pdfoutput=1

\usepackage[verbose=true,letterpaper]{geometry} 
\AtBeginDocument{
    \newgeometry{
        textheight=9in,
        textwidth=5.5in,
        top=1in,
        headheight=12pt,
        headsep=25pt,
        footskip=30pt
    }
}

\usepackage[utf8]{inputenc} 
\usepackage[T1]{fontenc}    
\usepackage{amsfonts}       
\usepackage{amssymb}
\usepackage{amsmath}
\usepackage{amsthm}
\usepackage{nicefrac}       
\usepackage{hyperref}       
\usepackage{url}            
\usepackage{booktabs}       
\usepackage{microtype}      
\usepackage{wrapfig}
\usepackage{multirow}
\usepackage{float}          
\usepackage{doi}
\usepackage{minitoc}        
\usepackage{subcaption}
\usepackage{adjustbox}
\usepackage[font=small,labelfont=bf]{caption}
\usepackage[toc,page,header]{appendix}

\usepackage{xcolor}         
\usepackage{graphicx} 

\theoremstyle{definition}

\newtheorem{rem}{Remark}[section]

\definecolor{ggreen}{RGB}{17,119,51}
\definecolor{gred}{RGB}{204,102,119}
\definecolor{gblue}{RGB}{51,34,136}

\hypersetup{
    colorlinks=true,
    linkcolor=gblue,            
    citecolor=gblue,            
    filecolor=blue,             
    urlcolor=ggreen             
}

\input{commands_math.tex}

\newcommand{\ie}{{i.e.}}
\newcommand{\eg}{{e.g.}}
\newcommand{\etal}{{et al.\ }}
\DeclareMathOperator{\vectorize}{vec}

\DeclareMathOperator{\softmax}{softmax}

\title{Topological Attention for Time Series Forecasting}

\author{%
    \textbf{Sebastian Zeng}\\
    Department of Computer Science\\
    University of Salzburg\\
    \texttt{sebastian.zeng@sbg.ac.at} \\
    \And 
    \textbf{Florian Graf}\\
    Department of Computer Science\\
    University of Salzburg\\
    \texttt{florian.graf@sbg.ac.at} \\
    \And
    \textbf{Christoph Hofer}\\
    Department of Computer Science\\
    University of Salzburg\\
    \texttt{chofer@cosy.sbg.ac.at} \\
    \And 
    \textbf{Roland Kwitt}\\
    Department of Computer Science\\
    University of Salzburg\\
    \texttt{roland.kwitt@sbg.ac.at} \\
}

\hypersetup{
    pdftitle={Topological Attention for Time Serie Forecasting},
    pdfkeywords={Time series forecasting, Persistent homology, Attention, Topological Data Analysis}}

\begin{document}

\doparttoc
\faketableofcontents

\pagecolor{white}
\maketitle

\begin{abstract}
    The problem of (point) forecasting \emph{univariate} time series is considered.
    Most approaches, ranging from traditional statistical methods to recent
    learning-based techniques with neural networks, directly operate on raw time
    series observations. As an extension, we study whether \emph{local topological
        properties}, as captured via persistent homology, can serve as a reliable signal
    that provides complementary information for learning to forecast. To this end,
    we propose \emph{topological attention}, which allows attending to local
    topological features within a time horizon of historical data. Our approach
    easily integrates into existing end-to-end trainable forecasting models, such as
    \texttt{N-BEATS}, and in combination with the latter exhibits state-of-the-art
    performance on the large-scale M4 benchmark dataset of 100,000 diverse time
    series from different domains. Ablation experiments, as well as a comparison to
    a broad range of forecasting methods in a setting where only a single time
    series is available for training, corroborate the beneficial nature of including
    local topological information through an attention mechanism.
\end{abstract}


\vspace{0.3cm}
\section{Introduction}
\label{section:introduction}
\input{sec_introduction}

\newpage

\section{Related work}
\label{section:sec_relatedwork}
\input{sec_relatedwork}

\vspace{-0.25cm}
\section{Topological attention}
\label{section:topological_attention}
\input{sec_topological_attention}

\section{Experiments}
\label{section:experiments}
\input{sec_experiments}

\section{Conclusion}
\label{section:conclusion}
\input{sec_conclusion}

\bibliographystyle{plain}
\bibliography{refs.bib, libraryChecked.bib, booksLibrary.bib}

\clearpage 

\appendix
\numberwithin{table}{section} 
\renewcommand{\thetable}{\Alph{section}.\arabic{table}} 

\par\noindent\rule{\textwidth}{1.2pt}
\begin{center}
     \textcolor{gblue}{\huge Supplementary Material}
\end{center}
\vskip -0.18cm
\par\noindent\rule{\textwidth}{0.4pt}

\counterwithin{figure}{section}

\mtcsetrules{parttoc}{off}
\mtcsettitle{parttoc}{Contents}
\mtcsetfont{parttoc}{section}{\rmfamily\upshape\mdseries}

\emph{This supplementary material contains additional results for the main
manuscript, a description and descriptive statistics of the datasets, as well as
training/architecture details for all models we compare to. When referring to
figures/tables/sections/etc. from the main manuscript, we use
\textbf{[Manuscript, Figure/Table/Section~X]} to identify the corresponding
parts.}

\vspace{-1.5cm}
\part{} 
\parttoc 

\clearpage 

\label{section:appendix}
\input{sec_appendix}

\end{document}

%% file: commands_math.tex
\newcommand{\bbR}{\mathbb{R}}

\newcommand{\mcB}{\mathcal{B}}

\newcommand{\mcK}{\mathcal{K}}

\newcommand{\bsyx}{\boldsymbol{x}}

\newcommand\restr[2]{{
  \left.\kern-\nulldelimiterspace 
  #1 
  \vphantom{\big|} 
  \right|_{#2} 
  }}


%% file: sec_introduction.tex
Time series are ubiquitous in science and industry, from medical signals (e.g., EEG), motion data (e.g., speed, steps, etc.) or economic operating figures to ride/demand volumes of transportation network companies (e.g., Uber, Lyft, etc.). Despite many advances in predicting future observations from historical data via traditional statistical \cite{Brockwell91a}, or recent machine-learning approaches \cite{Oreshkin20a,Wang19a,Salinas19a,Rangapuram18a,Li19a,Yu16a}, reliable and accurate forecasting remains challenging. This is not least due to widely different and often heavily domain dependent structural properties of time related sequential data.

In this work, we focus on the problem of (point) forecasting \emph{univariate} time series, i.e., given a length-$T$ vector of historical data, the task is to predict future observations for a given time horizon $H$.
While neural network models excel in situations where a large corpus of time series is available for training, the case of only a single (possibly long) time series is equally important. The arguably most prominent benchmarks for the former type of forecasting problem are the ``(M)akridakis''-competitions, such as M3 \cite{Makridakis00a} or M4 \cite{Makridakis18a}. While combinations (and hybrids) of statistical and machine-learning approaches have largely dominated these competitions \cite[see Table 4]{Makridakis18a}, Oreshkin et al. \cite{Oreshkin20a} have recently demonstrated that a pure machine-learning based model (\texttt{N-BEATS}) attains state-of-the-art performance. Interestingly, the latter approach is simply built from a collection of common neural network primitives which are \emph{not} specific to sequential data. 

However, the majority of learning-based approaches directly operate on the raw input signal, implicitly assuming that viable representations for forecasting can be learned via common neural network primitives, composed either in a feedforward or recurrent manner. This raises the question of whether there exist structural properties of the signal, which are not easily extractable via neural network components, but offer complementary information. 
One prime example\footnote{Although learning-based approaches \cite{Som20a} exist to approximate topological summaries (without guarantees).} are \emph{topological features}, typically obtained via persistent homology \cite{Carlsson09a,Edelsbrunner2010}. In fact, various approaches \cite{Perea15a,Gidea18a,Dlotko19a,Khasawneh18a,Gidea20a} have successfully used topological features for time series analysis, however, mostly in classification settings, for the identification of certain phenomena in dynamical systems, or for purely exploratory analysis (see Section~\ref{section:sec_relatedwork}).

\textbf{Contribution.} 
We propose an approach to incorporate \emph{local} topological information into neural forecasting models. Contrary to previous works, we do not compute a global topological summary of historical observations, but features of short, overlapping time windows to which the forecasting model can attend to. The latter is achieved via self-attention and thereby integrates well into recent techniques, such as \texttt{N-BEATS} \cite{Oreshkin20a}. Notably, in our setting, computation of topological features (via persistent homology) comes with little computational cost, which allows application in large-scale forecasting problems.

\begin{figure}
\includegraphics[width=0.98\textwidth]{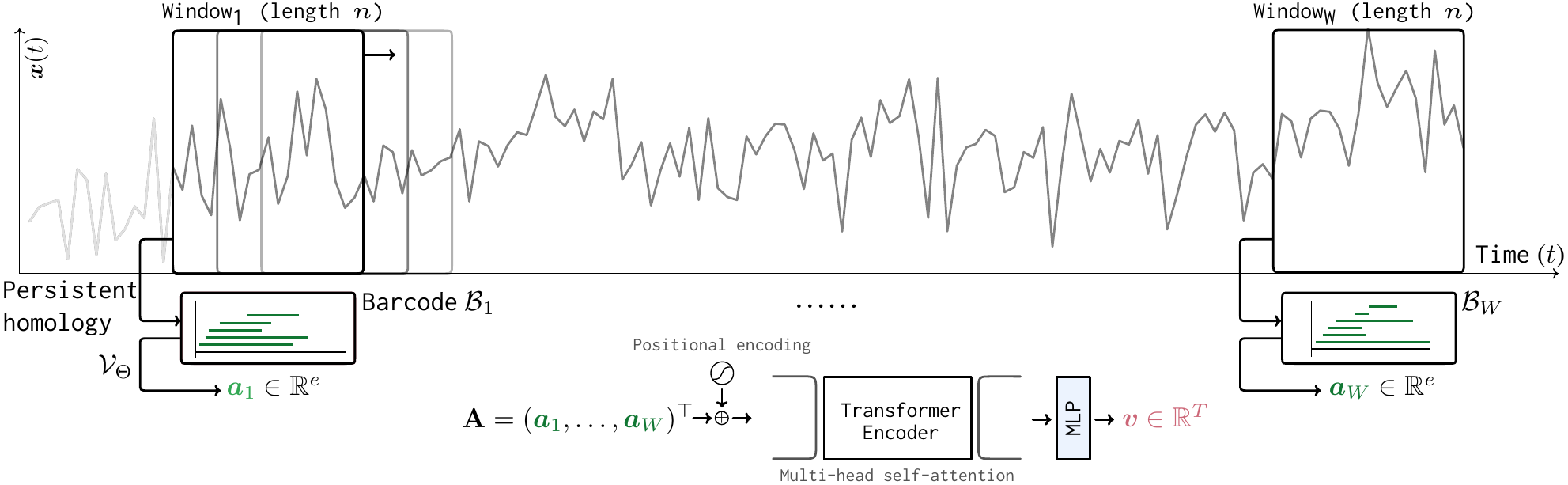}
\caption{Illustration of \emph{topological attention}, computed on time series observations $x_1,\ldots,x_T$. 
The signal is  decomposed into a collection of $W$ overlapping windows of length $n$. For each window, a topological summary, \ie, a persistence barcode $\mathcal{B}_j$, is computed. 
These \emph{local} topological summaries are then vectorized (in $\mathbb{R}^e$) via a differentiable map $\mathcal{V}_\Theta$, fed through several transformer encoder layers \cite{Vaswani17a} (implementing a multi-head self-attention mechanism) with positional encoding at the input, and finally mapped to $\boldsymbol{v} \in \mathbb{R}^T$ by an MLP \label{fig:intro} (best-viewed in color).}
\end{figure}

\textbf{Problem statement.}
In practice, neural forecasting models typically utilize the last $T$ observations of a time series in order to yield (point) forecasts for a given time horizon $H$.
Under this perspective, the problem boils down to learning a function (parametrized as a neural network) 
\begin{equation}
\phi: \mathbb{R}^T \to \mathbb{R}^H\quad \boldsymbol{x} \mapsto \phi(\bsyx) = \boldsymbol{y}\enspace,
\end{equation}
from a given collection of inputs (\ie, length-$T$ vectors) and targets (\ie, length-$H$ vectors). 
Specifically, we consider two settings, where either (1) a large collection of time series is available, as in the M4 competition, or (2) we only have access to a single time series. In the latter setting, a model has to learn from patterns \emph{within} a time series, while the former setting allows to exploit common patterns \emph{across} multiple time series.

%% file: sec_relatedwork.tex
\textbf{Persistent homology and time series.} Most  approaches to topological time series analysis 

are conceptually similar, building on top of work by de Silva et al. \cite{deSilva12a} and Perea \& Harer \cite{Perea15a,Perea16a}. 
Herein, time series observations are transformed into a point cloud via a \emph{time-delay coordinate embedding}  \cite{Takens81a} from which \emph{Vietoris-Rips (VR) persistent homology} is computed. The resulting topological summaries, \ie, persistence barcodes, are then used for downstream processing. 
Within this regime, Gidea et al. \cite{Gidea20a} analyze the dynamics of cryptocurrencies using persistence landscapes \cite{Bubenik15a}, Khasawneh et al. \cite{Khasawneh18a} study chatter classification in synthetic time series from turning processes and D\l{}otko et al. \cite{Dlotko19a} identify periodicity patterns in time series. 
In \cite{Kim18a}, Kim et al. actually compute one-step forecasts for Bitcoin prices and classify price patterns, essentially feeding barcode statistics as supplementary features to a MLP/CNN-based regression model. 
Surprisingly, very few works deviate from this pipeline, with the notable exception of \cite{Gidea18a}, where VR persistent homology is  \emph{not} computed from a time-delay coordinate embedding, but rather from assembling observations (within sliding windows of size $n$) from a $d$-variate time series into a $d$-dimensional point cloud, followed by VR persistent homology computation.

Although these works clearly demonstrate that capturing the ``shape'' of data via persistent homology provides valuable information for time series related problems, they (1) rely on \emph{handcrafted} features (i.e., predefined barcode summary statistics, or a fixed barcode-vectorization strategy), (2) consider topological summaries as the \emph{single source} of information and (3) only partially touch upon forecasting problems (with the exception of \cite{Kim18a}). 
Furthermore, in this existing line of work, sweeping a sliding window over the time series is, first and foremost, a way to construct a point cloud which represents the \emph{entire} time series.  
 Instead, in our approach, each window yields its \emph{own} topological summary in the form of a \emph{persistence barcode}, reminiscent to representing a sentence as a sequence of word embeddings in NLP tasks. 
 When combined with learnable representations of persistence barcodes  \cite{Hofer19b,Carriere20a}, this perspective paves the way for leveraging recent techniques for handling learning problems with sequential data, such as attention \cite{Vaswani17a}, and allows to seamlessly integrate topological features into existing neural  forecasting techniques.

\textbf{Neural network approaches to time series forecasting.} 
Recently, various  successful neural network approaches to (mostly probabilistic) time series forecasting have emerged, ranging from auto-regressive neural networks as in DeepAR \cite{Salinas19a}, to (deep) extensions of traditional state space models, such as DeepFactors \cite{Wang19a} or DeepState \cite{Rangapuram18a}.
While these models are inherently tailored to the sequential nature of the forecasting problem, Li et al. \cite{Li19a} instead rely on the concept of (log-sparse) self-attention \cite{Vaswani17a}, fed by the outputs of causal convolutions, and Oreshkin et al. \cite{Oreshkin20a} even abandon sequential neural network primitives altogether. 
The latter approach, solely based on operations predominantly found in feed-forward architectures, achieves state-of-the-art performance for (point) forecasts across several benchmarks, including the large-scale M4 competition. 
Yet, a common factor in all aforementioned works is that raw time series observations are directly input to the model, assuming that relevant structural characteristics of the signal can be learned. 
While we choose an approach similar to \cite{Li19a}, in the sense that we rely on self-attention, our work differs in that representations fed to the attention mechanism are not obtained through convolutions, but rather through a topological analysis step which, by its construction, captures the ``shape'' of local time series segments.

%% file: sec_topological_attention.tex
The key idea of topological attention is to analyze \emph{local} segments within an input time series, $\boldsymbol{x}$, through the lens of persistent homology. As mentioned in Section~\ref{section:sec_relatedwork}, the prevalent strategy in prior work is, to first construct a point cloud from $\boldsymbol{x}$ via a time-delay coordinate embedding and to subsequently compute VR persistent homology. 
Historically, this is motivated by studying structural properties of an underlying dynamical system, with a solid theoretical foundation, \eg, in the context of identifying periodicity patterns \cite{Perea15a,Perea16a}. 
In this regime, $\boldsymbol{x}$ is encoded as a point cloud in $\mathbb{R}^n$ by considering observations within a sliding window of size $n$ as a point in $\mathbb{R}^n$. 

While the time-delay embedding strategy is adequate in settings where one aims to obtain \emph{one global} topological summary, it is inadequate when local structural properties of time series segments are of interest. Further, unless large (computationally impractical) historical time horizons are considered, one would obtain relatively sparse point clouds that, most likely, carry little information.

\subsection{Time series as local topological summaries} 
Different to time-delay embeddings, we follow an alternative strategy: a time series signal $\boldsymbol{x}$ is still decomposed into a sequence of (overlapping) windows, but not to yield a point cloud element, but rather to be analyzed in \emph{isolation}. 
In the following, we only discuss the necessities specific to our approach, and refer the reader to \cite{Edelsbrunner02a,Carlsson09a,Boissonnat18a} for a thorough treatment of persistent homology.

To topologically analyze a length-$T$ time series, over the time steps $\{1, \dots, T\}=[T]$, in a computationally tractable manner, lets consider a 1-dimensional simplicial complex of the form
\[
  \mathcal{K} = \big\{ \{1\}, \dots \{T\}, \{1, 2\}, \dots, \{{T-1}, {T}\}\big\}
  \enspace,
\]
where $\{i\}$ denote 0-simplices (\ie, vertices) and 1-simplices $\{i,j\}$ (\ie, edges) are in $\mathcal{K}$ iff $i$ and $j$ are two consecutive time indices.
Topologically, $\mathcal{K}$ carries the connectivity properties of a time series of length $T$, which is equivalent to a straight line. This is the same for all time series of length $T$ and thus offers no discriminative information.

Persistent homology, however, lets us combine the purely topological representation of the time series with its actual values. For a specific $\boldsymbol{x}$, let $a_1 \leq \dots \leq a_T$ denote its increasingly sorted values and consider 
\[
  \mcK_{\boldsymbol{x}}^0 = \emptyset,  
  \quad\quad
  \mathcal{K}_{\boldsymbol{x}}^j = \{\sigma \in \mathcal{K}: \forall i \in \sigma: x_i \leq a_j\}
  \quad\text{for}\quad j \in [T]\enspace.
\]

In fact, $\emptyset = \mcK_{\boldsymbol{x}}^0 \subseteq \mathcal{K}_{\boldsymbol{x}}^1 \subseteq \dots \subseteq\mathcal{K}_{\boldsymbol{x}}^T = \mcK$ forms an increasing sequence of subsets of $\mathcal{K}$, \ie, a \emph{filtration}.
Importantly, while $\mcK$ is the same for all time series of length $T$, the filtration, $(\mcK_{\boldsymbol{x}}^j)_{j=0}^T$, is determined by the \emph{values} of $\boldsymbol{x}$.
Persistent homology then tracks the evolution of topological features throughout this sequence and summarizes this information in the form of \emph{persistence barcodes}. 

\begin{figure}[t]
  \includegraphics[width=0.98\textwidth]{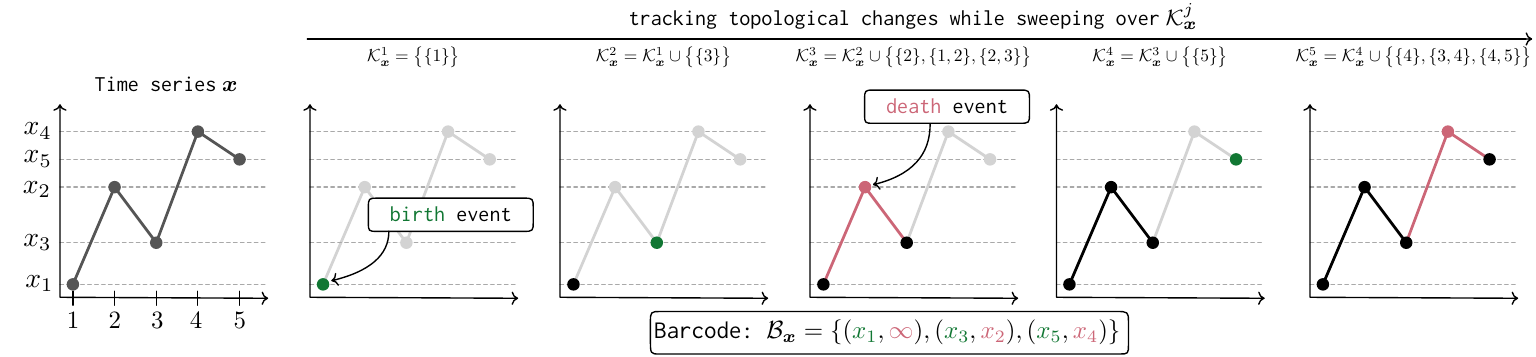}
  \caption{\label{fig:lowerstar} Illustration of $0$-dimensional persistent homology computation for a time series $\boldsymbol{x}$ of length $T=5$. The barcode $\mathcal{B}_{\boldsymbol{x}}$ encodes topological changes, in the form of (birth,death) tuples, as we sweep through the growing sequence $\mathcal{K}_{\boldsymbol{x}}^1 \subseteq \dots \subseteq\mathcal{K}_{\boldsymbol{x}}^T$ of subsets of $\mathcal{K}$. For example, the connected component \emph{born} at $x_3$,  \emph{dies} at $x_2$, caused by the merge with the connected component born at $x_1$ (best-viewed in color).}
\end{figure}

In our specific case, as $\mcK$ is topologically equivalent to a straight line, we only get $0$-dimensional features, \ie, connected components. Hence, we obtain one (0 degree) barcode $\mathcal{B}_{\boldsymbol{x}}$. This barcode is a multiset of (birth, death) tuples, representing the birth ($b$) and death ($d$) of topological features. Informally, we may think of building $\mathcal{K}$, piece-by-piece, according to the sorting of the $x_i$'s, starting with the lowest value, and tracking how connected components \emph{appear} / \emph{merge}, illustrated in Fig.~\ref{fig:lowerstar}. 

\begin{rem}
  \label{rem:phproperties}
The information captured throughout this process has two noteworthy properties. 
First, it is \emph{stable} in the sense that small changes in the observation values may not cause arbitrary changes in the respective barcodes, see \cite{Cohen-Steiner2007}. Second, one may equally order the negative observations, \ie, $-\boldsymbol{x}$, and thus obtain $\mathcal{B}_{-\boldsymbol{x}}$. In that manner, the signal is analyzed from \emph{below} and \emph{above}.
\end{rem}

Finally, to extract \emph{local topological information}, we do not compute one single barcode for $\boldsymbol{x}$, but one for each sliding window of size $n$, see Fig.~\ref{fig:intro}. Given a decomposition of $\boldsymbol{x}$ into $W$ subsequent windows, we obtain $W$ barcodes, $\mathcal{B}_1,\ldots,\mathcal{B}_W$, which constitute the entry point for any downstream operation. \emph{Informally, those barcodes encode the evolution of local topological features over time.}

\subsection{Barcode vectorization} 
Although persistence barcodes concisely encode topological features, the space of persistence barcodes, denoted as $\mathbb{B}$, carries no linear structure \cite{Turner14a} and the nature of barcodes as multisets renders them difficult to use in learning settings. Myriad approaches have been proposed to alleviate this issue, ranging from fixed mappings into a vector space (\eg, \cite{Bubenik15a,Adams17a}), to kernel techniques (\eg, \cite{Reininghaus14a,Kusano16a}) and, more recently, to learnable vectorization schemes (\eg, \cite{Hofer19a,Carriere20a}). Here, we follow the latter approach, as it integrates well into the regime of neural networks. In particular, the core element in learnable vectorization schemes is a differentiable map of the form 
\begin{equation}
  \mathcal{V}_\theta: \mathbb{B} \to \mathbb{R}, \quad \mathcal{B} \mapsto \sum_{(b,d) \in \mathcal{B}} s_\theta(b,d)\enspace, 
  \label{eqn:vectorizationmap}
\end{equation}
where $s_\theta: \mathbb{R}^2 \to \mathbb{R}$ denotes a so called \emph{barcode coordinate function} \cite{Hofer19a}, designed to preserve the \emph{stability} property in Remark~\ref{rem:phproperties}. Upon assembling a collection of $e \in \mathbb{N}$ such coordinate functions and subsuming parameters into $\Theta$, one obtains a $e$-dimensional vectorization of $\mathcal{B} \in \mathbb{B}$ via
\begin{equation}
  \label{eqn:VTheta}
\mathcal{V}_{\Theta}: \mathbb{B} \to \mathbb{R}^e, \quad \mcB \mapsto \mathbf{a} = \big(\mathcal{V}_{\theta_1}(\mathcal{B}),\ldots,\mathcal{V}_{\theta_e}(\mathcal{B})\big)^\top\enspace.
\end{equation}
Taking into account the representation of $\boldsymbol{x}$ as $W$ persistence barcodes, we summarize the vectorization step as 
\begin{equation}
  \texttt{TopVec}: \mathbb{B}^W \to \mathbb{R}^{W \times e}, 
  \quad
  (\mathcal{B}_1,\ldots,\mathcal{B}_W) \mapsto (\boldsymbol{a}_1,\ldots,\boldsymbol{a}_W)^
  \top = \big(\mathcal{V}_\Theta(\mathcal{B}_1), \ldots,\mathcal{V}_\Theta(\mathcal{B}_W)\big)^\top\enspace.
  \label{eqn:topenc}
\end{equation}

This is distinctly different to \cite{Perea15a,Perea16a,Khasawneh18a,Kim18a} (see Section~\ref{section:sec_relatedwork}), where \emph{one} barcode is obtained and this barcode is represented in a \emph{fixed} manner, \eg, via persistence landscapes \cite{Bubenik15a} or via barcode statistics. 

\subsection{Attention mechanism} 
In order to allow a forecasting model to attend to local topological patterns, as encoded via the $\boldsymbol{a}_j$, we propose to use the encoder part of Vaswani et al.'s \cite{Vaswani17a} transformer architecture, implementing a repeated application of a (multi-head) self-attention mechanism. Allowing to attend to local time series segments is conceptually similar to Li et al. \cite{Li19a}, but differs in the way local structural properties are captured: not via causal convolutions, but rather through the lens of persistent homology.
In this setting, the scaled dot-product attention, at the heart of a transformer encoder layer, computes 

\begin{equation}
  \label{eqn:self_attention}
\mathbf{O} = 
\softmax \left( \frac{(\mathbf{A}\mathbf{W}^q)(\mathbf{A}\mathbf{W}^k)^\top}{\sqrt{d_k}} \right)\mathbf{A}\mathbf{W}^v\enspace, \quad \text{with} \quad \mathbf{A} \stackrel{\text{Eq.~\eqref{eqn:topenc}}}{=} (\boldsymbol{a}_1,\ldots,\boldsymbol{a}_W)^
\top\enspace, 
\end{equation}

and $\mathbf{W}^q \in \mathbb{R}^{e \times d_q}, \mathbf{W}^k \in \mathbb{R}^{e \times d_k}$,  $\mathbf{W}^v \in \mathbb{R}^{e \times d_v}$ denoting learnable (key, value, query) projection matrices. Recall that $\mathbf{A} \in \mathbb{R}^{W \times e}$ holds all $e$-dimensional vectorizations the $W$ persistence barcodes.
In its actual incarnation, one transformer encoder layer\footnote{omitting the additional normalization and projection layers for brevity}, denoted as $\texttt{AttnEnc}:\mathbb{R}^{W \times e} \to \mathbb{R}^{W \times e}$, computes and concatenates $M$ parallel instances of Eq.~\eqref{eqn:self_attention}, (\ie, the attention heads), and internally adjusts $d_v$ such that $d=Md_v$. 
Composing $E$ such \texttt{AttnEnc} maps, one obtains

\begin{equation}
\label{eqn:transformerencoder}
\begin{split}
\texttt{TransformerEncoder}: & ~\mathbb{R}^{W \times e} \to \mathbb{R}^{W \times e}\\ 
& ~\mathbf{A} \mapsto \texttt{AttnEnc}_1 \circ \cdots \circ \texttt{AttnEnc}_E(\mathbf{A})\enspace.
\end{split}
\end{equation}

Finally, we use a two-layer $\texttt{MLP}: \mathbb{R}^{We} \to \mathbb{R}^T$ (with ReLU activations) to map the vectorized output of the transformer encoder to a $T$-dimensional representation. \emph{Topological attention} thus implements
\begin{equation}
  \begin{split}  
\texttt{TopAttn}: & ~\mathbb{B}^W \to \mathbb{R}^T \\
                  & ~(\mathcal{B}_1,\ldots,\mathcal{B}_W) \mapsto \texttt{MLP}(\vectorize (\texttt{TransformerEncoder} \circ  \texttt{TopVec}(\mathcal{B}_1,\ldots,\mathcal{B}_W)))\enspace,
\end{split}
\label{eqn:topattn}
\end{equation}
where $\vectorize(\cdot)$ denotes row-major vectorization operation.

\begin{rem}
  \label{rem:bplimitations}
  Notably, as the domain of $\texttt{TopAttn}$ is $\mathbb{B}^W$, error backpropagation stops at the persistent homology computation. However, we remark that, given recent works on differentiating through the persistent homology computation \cite{Hofer20a,Carriere21a}, one could even combine topological attention with, \eg, \cite{Li19a}, in the sense that the outputs of causal convolutions could serve as the \emph{filter} function for persistent homology. Error backpropagation would then consequently allow to learn this filter function. 
\end{rem}

\subsection{Forecasting model}
\label{subsection:forecasting_model}

While the representation yielded by topological attention (see Eq.~\eqref{eqn:topattn}) can be integrated into different neural forecasting approaches, it specifically integrates well into the \texttt{N-BEATS} model of Oreshkin \etal \cite{Oreshkin20a}. We briefly describe the \emph{generic} incarnation of \texttt{N-BEATS} next, but remark that topological attention can be similarly integrated into the basis-expansion variant without modifications.  

\begin{wrapfigure}{r}{0.3\textwidth}
    \vspace{-8pt}
    \begin{center}
      \includegraphics[width=0.26\textwidth]{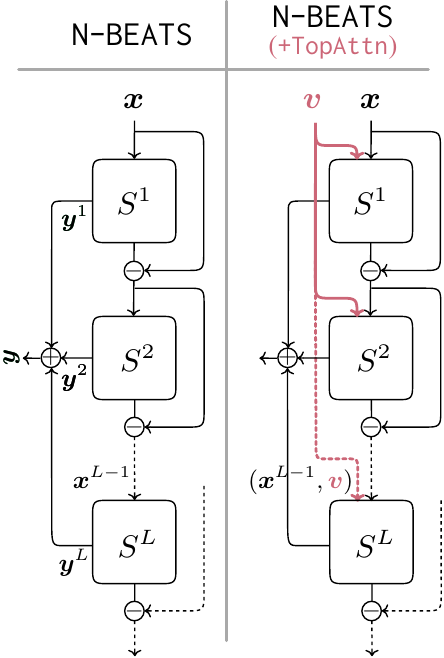}
    \end{center}
    \vspace{-15pt}
\end{wrapfigure}

Essentially, the generic \texttt{N-BEATS} model is assembled from a stack of $L$ double-residual blocks.
For $1 \leq l \leq L$,
each block consists of a non-linear map 
(implemented as a MLP)
\begin{equation}
S^l: \mathbb{R}^T \rightarrow \mathbb{R}^h\enspace,
\label{eqn:nbeatsblock}
\end{equation}
with $h$ denoting the internal dimensionality, and two subsequent 
maps that yield the two-fold output of the $l$-th block as
\begin{align*}
  \boldsymbol{x}^l = \boldsymbol{x}^{l-1} - \mathbf{U}^l \big(S^l(\boldsymbol{x}^{l-1})\big), \quad \text{and} \quad 
  \boldsymbol{y}^l = \mathbf{V}^l\big(S^l(\boldsymbol{x}^{l-1})\big)\enspace,
\end{align*}
with $\mathbf{U}^l \in \mathbb{R}^{T \times h}$, $\mathbf{V}^l \in \mathbb{R}^{H \times h}$ and $\boldsymbol{x}^0 = \boldsymbol{x}$. 
While the $\boldsymbol{x}^l$ yield the connection to the following computation block, the $\boldsymbol{y}^l$ are used to compute the final model prediction $\boldsymbol y\in \bbR^H$ via (component-wise) summation, \ie, $\boldsymbol{y} = \boldsymbol{y}^1 + \cdots + \boldsymbol{y}^L$. 

Importantly, in this computational chain, the $\boldsymbol{x}^l$ can be leveraged as an interface to integrate additional information. In our case, we enrich the input signal to each block by the output of the topological attention mechanism through concatenation (see figure to the right), \ie, 
\begin{equation}
  \label{eqn:attnsignalinclusion}
  \boldsymbol{x}_{\texttt{TopAttn}}^l = (\boldsymbol{x}^l, \boldsymbol{v}) \quad 
  \text{where} \quad 
  \boldsymbol{v} = \texttt{TopAttn}(\mathcal{B}_1,\ldots,\mathcal{B}_W)\enspace.
\end{equation}

This means that the time series signal $\boldsymbol{x}$ is (1) input (in its raw form) to \texttt{N-BEATS} and (2) its topological attention representation, $\boldsymbol{v}$, is supplied to each block as a complementary signal. In a similar manner (\ie, through concatenation), $\boldsymbol{v}$ can be included in much simpler models as well (see Section~\ref{subsection:single_time_series_experiments}).

\textbf{Computational complexity.}
Aside from the computational overhead incurred by the multi-head attention module, we need to compute $0$-dimensional persistent homology for each sliding window.  
This can be done efficiently, using union find data structures, with complexity $\mathcal{O}\big(m \alpha^{-1}(m)\big)$, where 
$m=|\mathcal{K}|=2n-1$
with $n$ the sliding window size and $\alpha^{-1}(\cdot)$ denoting the inverse of the Ackermann function. As the latter grows \emph{very} slowly, computational complexity is roughly linear for this part.

%% file: sec_experiments.tex
We assess the  quality of \emph{point forecasts} in two different settings and perform ablation studies to isolate the impact of topological attention in the proposed regime. 

Throughout all experiments, we compute persistent homology from $\boldsymbol{x}$ \emph{and} $-\boldsymbol{x}$ (see Remark~\ref{rem:phproperties}) using \texttt{Ripser} \cite{Bauer21}. Barcode vectorization, see Eq.~\eqref{eqn:VTheta}, is based on {rational hat} coordinate functions \cite{Hofer19a} with the position parameters (\ie, the locations of each coordinate function in $\mathbb{R}^2$) initialized by $k$-means++ clustering over all barcodes in the training data (with $k$ set to the number of coordinate functions). This yields a representation $\boldsymbol{a} \in \mathbb{R}^{2e}$ per sliding window. \emph{Full architecture details and dataset statistics can be found in the suppl. material}.

\textbf{Ablation setup.} When assessing each component of topological attention in isolation, we refer to \texttt{+Top} as omitting the \texttt{TransformerEncoder} part in Eq.~\eqref{eqn:topattn}, and to \texttt{+Attn} as directly feeding the time series observations to the transformer encoder, \ie, omitting \texttt{TopVec} in Eq.~\eqref{eqn:topattn}.

\subsection{Evaluation metrics}
\label{subsection:evaluation_metrics}
 To evaluate the quality of point forecasts, two commonly used metrics are the \emph{symmetric mean absolute percentage error} (\texttt{sMAPE}) and the \emph{mean absolute scaled error} (\texttt{MASE}). Letting $\hat{\boldsymbol{y}}=(
    \hat{x}_{T+1},\ldots,\hat{x}_{T+H})^\top$ denote the length-$H$ forecast, $\boldsymbol{y}=(x_{T+1},\ldots,x_{T+H})^\top$ the true observations and $\boldsymbol{x} = (x_1,\ldots,x_T)^\top$ the length-$T$ history of input observations, both scores are defined as \cite{Makridakis20a}

\begin{equation}
    \label{eqn:smapemase}
    \text{\texttt{sMAPE}}(\boldsymbol{y},\hat{\boldsymbol{y}}) =
    \frac{200}{H} \sum_{i=1}^H \frac{|x_{T+i} - \hat{x}_{T+i}|}{|x_{T+i}| + |\hat{x}_{T+i}|},\
    \text{\texttt{MASE}}(\boldsymbol{y},\hat{\boldsymbol{y}}) = \frac{1}{H}\frac{\sum_{i=1}^{H}|x_{T+i} - \hat{x}_{T+i}|}{\frac{1}{T-m}\sum_{i=m+1}^T|x_i - x_{i-m}|}\enspace,
\end{equation}

with $m$ depending on the observation frequency. For results on the M4 benchmark (see Section~\ref{subsection:large_scale_experiments_on_the_M4_benchmark}), we adhere to the competition guidelines and additionally report the \emph{overall weighted average} (\texttt{OWA}) which denotes the arithmetic mean of \texttt{sMAPE} and \texttt{MASE} (with $m$ pre-specified), both measured relative to a na\"ive (seasonally adjusted) forecast (also provided by the M4 competition as \texttt{Naive2}).

\subsection{Single time series experiments}
\label{subsection:single_time_series_experiments}

We first consider the simple, yet frequently occurring, practical setting of \emph{one-step} forecasts with historical observations available for only a \emph{single} length-$N$ time series.
Upon receiving a time series $\boldsymbol{x} \in \mathbb{R}^T$ (with $T \ll N$), a model should yield a forecast for the time point $T+1$ (\ie, $H=1$).

\subsubsection{Dataset}
To experiment with several single (but long) time series of different characteristics, we use 10 time series from the publicly available \texttt{electricity} \cite{Dua17a} demand dataset\footnote{\url{https://archive.ics.uci.edu/ml/datasets/ElectricityLoadDiagrams20112014}} and four (third-party) time series of car part demands, denoted as \texttt{car-parts}. Based on the categorization scheme of \cite{Syntetos05a}, the time series are chosen such that not only \emph{smooth} time series (regular demand occurrence and low demand quantity variation) are represented, but also \emph{lumpy} ones (irregular demand occurrence and high demand quantity variation). For \texttt{electricity}, the respective percentages are 70\% vs. 30\%, and, for \texttt{car-parts},  75\% vs. 25\%. All observations are \emph{non-negative}. In case of \texttt{electricity}, which contain measurements in 15min intervals, we aggregate (by summation) within 7h windows, yielding a total of 3,762 observations.
For \texttt{car-parts}, demand is measured on daily basis across a time span of 7-8 years (weekends and holidays excluded), yielding 4,644 
observations on average. For each time series, 20\% of held-out consecutive observations are used for testing, 5\% for validation.

\subsubsection{Forecasting model} 
We employ a simple incarnation of the forecasting model from Section~\ref{subsection:forecasting_model}. In particular, we replace \texttt{N-BEATS} by a single linear map (with bias), implementing
\begin{equation}
    \big((\mathcal{B}_1,\ldots,\mathcal{B}_W),\boldsymbol{x}\big) \mapsto \boldsymbol{w}^\top 
    \boldsymbol{x}_{\text{\texttt{TopAttn}}} + b\enspace,
    \label{eqn:singlelinmap}
\end{equation}
with $\boldsymbol{x}_{\text{\texttt{TopAttn}}}$ denoting the concatenation of the topological attention signal and the input time series $\boldsymbol{x}$, as in Eq.~\eqref{eqn:attnsignalinclusion}. During training, we randomly extract $T+1$ consecutive observations from the training portion of the time series. The first $T$ observations are used as input $\boldsymbol{x}$, the observation at $T+1$ is used as target. Forecasts for all testing observations are obtained via a length-$T$ rolling window, moved forward one step a time. 

In terms of hyperparameters for topological attention, we use a single transformer encoder layer with four attention heads and 32 barcode coordinate functions. We minimize the mean-squared-error via ADAM over 1.5k (\texttt{electricity}) and 2k (\texttt{car-parts}) iterations, respectively, with a batch size of 30. Initial learning rates for the components of Eq.~\eqref{eqn:topattn} are 
9e-2 (\texttt{TopVec}, \texttt{MLP}) and 5e-2 (\texttt{TransformerEncoder}), as well as 9e-2 for the linear map of Eq~\eqref{eqn:singlelinmap}. All learning rates are annealed following a cosine learning rate schedule. 

\begin{table}[t!]
\begin{center}
    \begin{small}
    \caption{\label{table:single} Single time series experiments on \texttt{car-parts} and \texttt{electricity}, using the \texttt{sMAPE} as performance criterion. Listed are (1) the \emph{average rank} ($\oslash$ Rank) of each method, as well as (2) the \emph{average percentual difference} (\% Diff.) to the Rank-1 approach per time series. $\dagger$ denotes \texttt{GluonTS} \cite{gluonts_jmlr} implementations.}
\hspace{10pt}
\begin{subtable}[t]{0.48\textwidth}
\begin{center}
\caption{\texttt{car-parts (4)}}
\begin{tabular}{lcc}
\toprule
\textbf{Method} & $\oslash$  \textbf{Rank} & \% \textbf{Diff.} \\
\midrule 
\texttt{Lin.+\textcolor{gred}{TopAttn}}      & \textbf{1.50} & \textbf{2.82} \\
\texttt{Prophet}          & 2.75 & 4.91 \\
$^\dagger$\texttt{MLP}     & 3.00 & 7.49 \\
$^\dagger$\texttt{DeepAR}  & 3.50 & 7.86 \\
\texttt{autoARIMA}        & 5.25 & 13.45 \\
\texttt{LSTM}             & 6.00 & 16.24 \\
$^\dagger$\texttt{MQ-RNN}  & 7.50 & 34.36 \\
\texttt{Naive}            & 7.75 & 30.08 \\
$^\dagger$\texttt{MQ-CNN}  & 7.75 & 29.19 \\
\bottomrule    
\end{tabular}
\end{center}
\end{subtable}
\begin{subtable}[t]{0.48\textwidth}
    \begin{center}
\caption{\texttt{electricity (10)}}
\begin{tabular}{lcc}
\toprule

\textbf{Method} & $\oslash$  \textbf{Rank} & \% \textbf{Diff.} \\
\midrule 
\texttt{Lin.+\textcolor{gred}{TopAttn}}      & \textbf{1.50} & \textbf{10.59} \\
$^\dagger$\texttt{DeepAR}  & 1.90 & 12.14 \\
$^\dagger$\texttt{MLP}     & 2.90 & 15.39 \\
$^\dagger$\texttt{MQ-CNN}  & 4.70 & 45.44 \\
\texttt{autoARIMA}        & 5.10 & 45.67 \\
\texttt{Prophet}          & 6.10 & 61.55 \\
\texttt{LSTM}             & 7.40 & 69.89 \\
$^\dagger$\texttt{MQ-RNN}  & 7.50 & 68.74 \\
\texttt{Naive}            & 7.90 & 77.71 \\
\bottomrule    
\end{tabular}
\end{center}
\end{subtable}
\end{small}
\end{center}
\vspace{-5pt}
\end{table}

\subsubsection{Results \& Ablation study}

We compare against several techniques from the literature that are readily available to a practitioner. This includes \texttt{autoARIMA} \cite{Hyndman08b}, \texttt{Prophet} \cite{Taylor17a}, a vanilla \texttt{LSTM} model, as well as several approaches implemented within the \texttt{GluonTS} \cite{gluonts_jmlr} library. With respect to the latter, we list results for a single-hidden-layer \texttt{MLP}, \texttt{DeepAR} \cite{Salinas19a} and \texttt{MQ-CNN/MQ-RNN} \cite{Wen17a}. By \texttt{Naive}, we denote a baseline, yielding $x_{T}$ as forecast for $x_{T+1}$. \emph{Importantly, each model is fit separately to each time series in the dataset.}

For a fair comparison, we further account for the fact that forecasting models typically differ in their sensitivity to the length of the input observations, $\boldsymbol{x}$. To this end, we cross-validate $T$ (for all methods) using the \texttt{sMAPE} on the validation set. Cross-validation points are determined by the topological attention parameters $W$ and $n$, \ie, the number and lengths of the sliding windows. For $n$ ranging from 10 to 200 and $W$ ranging from 5 to 45, we obtain a wide range of input lengths, from 14 to 244. Instead of listing \emph{absolute} performance figures, we focus on the \emph{average rank}\footnote{the \texttt{sMAPE} determines the rank of a method per time series; these ranks are then averaged over all time series } within the cohort of methods, as well as the \emph{average percentual difference} to the best-ranking approach per time series.

Table~\ref{table:single} lists the overall statistics for \texttt{electricity} and \texttt{car-parts}. We observe that, while the overall ranking per dataset differs quite significantly, \texttt{Lin+TopAttn} consistently ranks well. Second, the average percentual difference to the best-ranking approach per time series is low, meaning that while \texttt{Lin+TopAttn} might not yield the most accurate forecasts on a specific time series, it still produces forecasts of comparable quality.

Table~\ref{tbl:singleablation} provides the same performance statistics for an ablation study of the topological attention components. Specifically, we combine the linear model of Eq.~\eqref{eqn:singlelinmap} with each component of topological attention in isolation. 

\begin{wraptable}{r}{5.2cm}
    \begin{center}
    \vspace{-14pt}
    \begin{small}
    \caption{Ablation study\label{tbl:singleablation}}
    \vspace{-2pt}
    \begin{tabular}{lcc}
        \toprule
        & $\oslash$ \textbf{Rank} & \% \textbf{Diff.} \\
        \toprule  
        \multicolumn{3}{l}{\texttt{car-parts (4)}} \\  
        \midrule  
    \texttt{Lin.}           & 2.75 & 0.15 \\
    \midrule 
    \texttt{~+Top}         & 3.50 & 1.45 \\  
    \texttt{~+Attn}        & \textbf{1.25} & 0.15 \\ 
    \texttt{~+\textcolor{gred}{TopAttn}}     & 2.50 & 0.64 \\  
    \midrule
    \multicolumn{3}{l}{\texttt{electricity (10)}}\\
    \midrule
    \texttt{Lin.}           & 2.40 & 5.95 \\
    \midrule 
    \texttt{~+Top}         & 3.40 & 12.20 \\  
    \texttt{~+Attn}        & 2.30 & 2.41 \\ 
    \texttt{~+\textcolor{gred}{TopAttn}}     & \textbf{1.90} & 2.10 \\  
    \bottomrule
    \end{tabular}
    \end{small}
\end{center}
    \end{wraptable}

Some observations are worth pointing out: 
\emph{First}, the linear model (\texttt{Lin.}) alone already performs surprisingly well. 
This can possibly be explained by the fact that the task only requires one-step forecasts, for which the historical length-$T$ observations (directly preceding the forecast point) are already quite informative. \emph{Second}, directly including topological features (\ie, \texttt{+Top}) has a confounding effect. We hypothesize that simply vectorizing local topological information from all sliding windows, \emph{without any focus}, obfuscates relevant information, rather than providing a reliable learning signal. This also highlights the importance of attention in this context,  which, even when \emph{directly} fed with observations from each sliding window (\ie, \texttt{+Attn}), exhibits favorable performance (particularly on \texttt{car-parts}). However, in the latter strategy, the input dimensionality for the transformer encoder scales with the sliding window size $n$.
Contrary to that, in case of topological attention, the input dimensionality is always fixed to the of number of coordinate functions, irrespective of the sliding window size $n$.

\subsection{Large-scale experiments on the M4 benchmark}
\label{subsection:large_scale_experiments_on_the_M4_benchmark}

Different to Section~\ref{subsection:single_time_series_experiments}, we now consider having \emph{multiple} time series of different lengths and characteristics available for training. Further, instead of one-step forecasts, the sought-for model needs to output (multi-step) point forecasts for time horizons $H>1$.

\subsubsection{Dataset} 
Experiments are based on the publicly available M4 competition dataset\footnote{available at \url{https://github.com/Mcompetitions/M4-methods}}, consisting of 100,000 time series from six diverse domains, aggregated into six \emph{subgroups} that are defined by the frequency of observations (\ie, yearly, quarterly, monthly, weekly, daily and hourly). Forecasting horizons range from $H=6$ (yearly) to $H=48$ (hourly). The test set is fixed and contains, for all time series in each subgroup, exactly $H$ observations to be predicted (starting at the last available training time point).

\subsubsection{Forecasting model} 
\label{subsection:forecasting_model_large_scale}
We employ the forecasting model\footnote{based on the \texttt{N-BEATS} reference implementation \url{https://github.com/ElementAI/N-BEATS}} of Section~\ref{subsection:forecasting_model} and largely stick to the architecture and training configuration of \cite[Table 18]{Oreshkin20a}. Our implementation only differs in the hidden dimensionality of \texttt{N-BEATS} blocks (128 instead of 512) and in the \emph{ensembling} step. In particular, for each forecast horizon (\ie, for each subgroup), \cite{Oreshkin20a} train multiple models, varying $T$ from $T=2H$ to $T=7H$, using ten random initializations and three separate loss functions (\texttt{sMAPE}, \texttt{MASE}, \texttt{MAPE}).
One final forecast per time series is obtained by median-aggregation of each model's predictions. In our setup, we solely rely on the \texttt{sMAPE} as loss function, vary $T$ only from $T=2H$ to $T=5H$, but still use ten random initializations. Even with this (smaller) ensemble size (40 models per subgroup, instead of 180), \texttt{N-BEATS} alone already outperforms the winner of M4 (see Table~\ref{table:m4}). As we are primarily interested in the effect of integrating topological attention, sacrificing absolute performance for a smaller ensemble size is incidental. 

In terms of hyperparameters for topological attention, the length ($n$) of sliding windows is set to $n=\lfloor 0.7 \cdot T \rfloor$, where $T$ varies per subgroup as specified above. The model uses 20 transformer encoder layers with two attention heads and 64 structure elements for barcode vectorization. For optimization, we use ADAM with initial learning rates of 1e-3 (for \texttt{N-BEATS} and the \texttt{MLP} part of Eq.~\eqref{eqn:topattn}), 8e-3 (\texttt{TopVec}) and 5e-3 (\texttt{TransformerEncoder}). All learning rates are annealed according to a cosine learning rate schedule over 5,000 iterations with a batch size of 1,024.

\begin{table}
    \begin{center}
\begin{small}
    \caption{Performance comparison on the M4 benchmark in terms of \texttt{sMAPE} / \texttt{OWA}, listed by subgroup. \texttt{N-BEATS} and \texttt{N-BEATS+TopAttn} denote an ensemble formed by training multiple models, varying $T$ from $2H$ to $5H$ and randomly initializing each model ten times (\ie, a total of 40 models per subgroup). Forecasts are obtained by taking the median over the point forecasts of all models. $\dagger$ denotes results from 
    \cite{Makridakis18b,Makridakis20a}. \label{table:m4}} 
\vskip4pt
\begin{tabular}{lccccc}
    \toprule
\multirow{2}{*}{\textbf{Method}} & \textbf{Yearly}
                        & \textbf{Quarterly} 
                        & \textbf{Monthly} 
                        & \textbf{Others} 
                        & \textbf{\textcolor{gblue}{Average}} \\ 
                        & (23k)           
                        & (24k)
                        & (48k) 
                        & (5k)
                        & \textcolor{gblue}{(100k)} \\
                        \midrule
                        $^\dagger$Winner M4 \cite{Smyl20a}   & 13.176 / 0.778 & 9.679 / 0.847  & 12.126 / 0.836 & 4.014 /0.920 & \textcolor{gblue}{11.374 /  0.821}\\
                        $^\dagger$Benchmark                  & 14.848 / 0.867 & 10.175 / 0.890 & 13.434 / 0.920 & 4.987 / 1.039 & \textcolor{gblue}{12.555 / 0.898}\\
                        $^\dagger$\texttt{Naive2}                   & 16.342 / 1.000 & 11.011 / 1.000 & 14.427 / 1.000 & 4.754 / 1.000 & \textcolor{gblue}{13.564 / 1.000} \\
                        \midrule
                        \texttt{N-BEATS} \cite{Oreshkin20a}& 
                            13.149 / 0.776    & 
                            \textbf{9.684}  / \textbf{0.845}    & 
                            12.054 / 0.829    & 
                            \textbf{3.789}  / \textbf{0.857}    & 
                            \textcolor{gblue}{11.324 / 0.814}    \\
                        \texttt{N-BEATS+\textcolor{gred}{\textbf{TopAttn}}}     & 
                            \textbf{13.063} / \textbf{0.771}     & 
                            9.687  / \textbf{0.845}     & 
                            \textbf{12.025} / \textbf{0.828}     & 
                            3.803  / 0.860     & 
                            \textbf{\textcolor{gblue}{11.291}} / \textbf{\textcolor{gblue}{0.811}}    \\
\bottomrule
\end{tabular}
\end{small}
\end{center}
\vspace{-2pt}
\end{table}

\vspace{-5pt}
\subsubsection{Results \& Ablation study} 
Table~\ref{table:m4} lists the \texttt{sMAPE} and \texttt{OWA} for the winner of the M4 competition \cite{Smyl20a}, as well as the \texttt{Naive2} baseline (with respect to which the \texttt{OWA} is computed) and the M4 benchmark approach, obtained as the arithmetic mean over simple, Holt, and damped exponential smoothing.

\begin{wraptable}{r}{4.7cm}
    \begin{center}
    \vspace{-12pt}
    \begin{small}
    \caption{Ablation study\label{table:m4ablation}}
    \vspace{-0.5cm}
    \begin{tabular}{lcc}\\\toprule  
    \textbf{Method}     & \texttt{sMAPE} & \texttt{OWA} \\\midrule
    \texttt{N-BEATS}     & 11.488 & 0.827 \\
    \midrule 
    \texttt{~+Top}      & 11.505 & 0.920 \\  
    \texttt{~+Attn}     & 11.492 & 0.826 \\ 
    \texttt{~+\textcolor{gred}{TopAttn}}  & \textbf{11.466} & \textbf{0.824} \\  
    \bottomrule
    \end{tabular}
    \end{small}
    \vspace{-8pt}
\end{center}
    \end{wraptable} 

In terms of the \texttt{OWA}, we see an overall 0.4\% improvement over \texttt{N-BEATS} and a 1.2\% improvement over the M4 winner \cite{Smyl20a}. In particular, topological attention performs well on the large yearly / monthly subgroups of 23k and 48k time series, respectively. While \texttt{OWA} scores are admittedly quite close, the differences are non-negligible, considering the large corpus of 100k time series. In fact, several methods in the official M4 ranking differ by an even smaller amount with respect to the \texttt{OWA} measure.

Similar to the ablation results of Section~\ref{subsection:single_time_series_experiments}, the ablation study in Table \ref{table:m4ablation} (conducted for $T=2H$ only) reveals the beneficial effect of topological attention, in particular, the beneficial nature of allowing to \emph{attend} to local topological features. Contrary to the ablation in Table~\ref{tbl:singleablation}, we see that in this large-scale setting, \emph{neither} topological features (\texttt{+Top}), \emph{nor} attention (\texttt{+Attn}) \emph{alone} yield any improvements over \texttt{N-BEATS}; when integrated separately into the \texttt{N-BEATS} model, both components even deteriorate performance in terms of the \texttt{sMAPE}.

%% file: sec_conclusion.tex
While several prior forecasting works have pointed out the relevance of local structural information within historical observations (\eg, \cite{Li19a}), it is typically left to the model to learn such features from data. Instead, we present a direct approach for capturing the ``shape'' of local time series segments via persistent homology. Different to the typical application of the latter in signal analysis, we capture the evolution of topological features over time, rather than a global summary, and allow a forecasting model to attend to these local features. The so obtained \emph{topological attention mechanism} yields a complementary learning signal that easily integrates into neural forecasting approaches. In combination with \texttt{N-BEATS} \cite{Oreshkin20a}, for instance, large-scale experiments on the M4 benchmark provide evidence that including topological attention indeed allows to obtain more accurate point forecasts. 

\textbf{Societal impact.} Due to the ubiquity of time series data, forecasting in general, certainly touches upon a variety of societally relevant and presumably sensible areas. As our work has potential impact in that sense, we perform large-scale experiments over a wide variety of time series from different domains, thereby obtaining a broad picture of the overall forecasting performance. 

%% file: sec_appendix.tex
\section{\textsc{Evaluation Criteria}}

In the following, we first replicate the definition of two commonly used, scale-free evaluation metrics, i.e., the \emph{symmetric mean absolute percentage error}
(\texttt{sMAPE}) and the \emph{mean absolute scaled error} (\texttt{MASE}), see \textbf{[Manuscript, Section~\ref{subsection:evaluation_metrics}]}. These scale-free metrics are standard in the practice
of forecasting and used across all experiments in the manuscript. Subsequently, we define the \emph{overall weighted average} (\texttt{OWA}), i.e., a M4 competition specific performance measure used to rank competition entries. We also provide a toy calculation example.

Letting
    $\hat{\boldsymbol{y}}=(\hat{x}_{T+1},\ldots,\hat{x}_{T+H})^\top$ denote the
    length-$H$ forecast, $\boldsymbol{y}=(x_{T+1},\ldots,x_{T+H})^\top$ the true
    observations and $\boldsymbol{x} = (x_1,\ldots,x_T)^\top$ the length-$T$
    history of input observations, both metrics are defined as
    \cite{Makridakis20a}
\begin{equation}
    \label{eqn:smapemase}
    \text{\texttt{sMAPE}}(\boldsymbol{y},\hat{\boldsymbol{y}}) =
    \frac{200}{H} \sum_{i=1}^H \frac{|x_{T+i} - \hat{x}_{T+i}|}{|x_{T+i}| + |\hat{x}_{T+i}|},\
    \text{\texttt{MASE}}(\boldsymbol{y},\hat{\boldsymbol{y}}) = \frac{1}{H}\frac{\sum_{i=1}^{H}|x_{T+i} - \hat{x}_{T+i}|}{\frac{1}{T-m}\sum_{i=m+1}^T|x_i - x_{i-m}|}\enspace,
\end{equation}
with $m$ depending on the observation frequency. In the M4 competition \cite{Makridakis20a}, the frequencies per subgroup are: 12 for monthly, four for quarterly, 24 for hourly and one for yearly / weekly / daily data. To obtain the \texttt{OWA} of a given forecast method, say \texttt{Forecast}, we compute \cite{Oreshkin20a}

\begin{equation}
    \label{eqn:owa}
    \text{\texttt{OWA}}_{\text{\texttt{Forecast}}} = \frac{1}{2} \left\lbrack \frac{\text{\texttt{sMAPE}}_{\text{\texttt{Forecast}}}}{\text{\texttt{sMAPE}}_{\text{\texttt{Naive2}}}} +
                                                    \frac{\text{\texttt{MASE}}_{\text{\texttt{Forecast}}}}{\text{\texttt{MASE}}_{\text{\texttt{Naive2}}}} \right\rbrack\enspace.
\end{equation}

Thus, if \texttt{Forecast} displays a \texttt{MASE} of 1.63 and a \texttt{sMAPE}
of 12.65\% across the 100k time series of M4, while \texttt{Naive2} displays a \texttt{MASE} of 1.91 and a \texttt{sMAPE} of 13.56\%, the relative \texttt{MASE} and \texttt{sMAPE} of \texttt{Forecast} would be 1.63/1.91
= 0.85 and 12.65/13.56 = 0.93, respectively, resulting in an \texttt{OWA} of (0.93
+ 0.85)/2 = 0.89. According to \cite{Makridakis20a}, this indicates that, on average, \texttt{Forecast} is about 11\%
more accurate than \texttt{Naive2}, taking into account both \texttt{sMAPE} and \texttt{MASE}.

\textbf{Performance criteria for single time series experiments.} In our single time series experiments of \textbf{[Manuscript, Section~\ref{subsection:single_time_series_experiments}]}, we use the \texttt{sMAPE} as an underlying evaluation measure and compute the following two statistics: first, the average rank ($\oslash$ \textbf{Rank}) of each method based on the rank on each single time series in \texttt{car-parts} and \texttt{electricity} (both datasets are treated separately); and, second, the average percentual difference (\% \textbf{Diff}) to the best approach per time series. A calculation example, for four \emph{hypothetical} models and three time series (TS$_0$, TS$_1$, TS$_2$), is listed in Table~\ref{table:appendix:singleexample}.

\begin{table}[h!]
    \caption{\label{table:appendix:singleexample}Example calculation of the \textcolor{ggreen}{\emph{average rank}} (\textcolor{ggreen}{$\oslash$ \textbf{Rank}}) and the \textcolor{gred}{\emph{average percentual difference}} (\textcolor{gred}{\% \textbf{Diff}}), as reported in [Manuscript, Table 1]. In this calculation example, \underline{lower} scores are \underline{better}. For instance, on TS$_1$, the rank of \texttt{Model C} is 3 and the percentual difference to the best performing model on TS$_1$ (i.e., \texttt{Model A}) is $(1.0 - 11.8/12.2)\times 100 = 3.28$.}
\vspace{3pt}
\begin{center}
\begin{tabular}{cccccc}
\toprule  
                 & TS$_0$   & TS$_1$   & TS$_2$  & \textcolor{ggreen}{$\oslash$~\textbf{Rank}} & \textcolor{gred}{\% \textbf{Diff}} \\
                \midrule
\texttt{Model A} & 14.4 (\textcolor{ggreen}{3}, \textcolor{gred}{9.66}) & 
                   11.8 (\textcolor{ggreen}{1}, \textcolor{gred}{0.00}) & 
                   10.5 (\textcolor{ggreen}{2}, \textcolor{gred}{3.81}) & 
                   \textcolor{ggreen}{\textbf{2.00}} &
                   \textcolor{gred}{\textbf{6.73}} \\
\texttt{Model B} & 14.3 (\textcolor{ggreen}{2}, \textcolor{gred}{8.39}) & 
                   12.1 (\textcolor{ggreen}{2}, \textcolor{gred}{2.48}) & 
                   10.8 (\textcolor{ggreen}{3}, \textcolor{gred}{6.48}) & 
                   \textcolor{ggreen}{\textbf{2.33}} & 
                   \textcolor{gred}{\textbf{5.78}} \\
\texttt{Model C} & 13.1 (\textcolor{ggreen}{1}, \textcolor{gred}{0.00}) & 
                   12.2 (\textcolor{ggreen}{3}, \textcolor{gred}{3.28}) & 
                   11.1 (\textcolor{ggreen}{4}, \textcolor{gred}{9.01}) & 
                   \textcolor{ggreen}{\textbf{2.67}} &
                   \textcolor{gred}{\textbf{6.14}} \\
\texttt{Model D} & 14.5 (\textcolor{ggreen}{4}, \textcolor{gred}{9.66}) & 
                   13.1 (\textcolor{ggreen}{4}, \textcolor{gred}{9.92}) & 
                   10.1 (\textcolor{ggreen}{1}, \textcolor{gred}{0.00}) & 
                   \textcolor{ggreen}{\textbf{3.00}} & 
                   \textcolor{gred}{\textbf{9.79}}\\
\bottomrule
\end{tabular}
\end{center}
\end{table}

\clearpage
\section{\textsc{Dataset Details}}

For completeness, Table~\ref{table:appendix:m4stat} replicates \cite[Table 2]{Oreshkin20a}, providing an overview of the key statistics for the M4 competition dataset. For all results listed in the main manuscript, the subgroups \emph{Weekly}, \emph{Daily} and \emph{Hourly} are aggregated into \textbf{Others}, accounting for 5,000 time series overall.

\begin{table}[H]
    \begin{center}
    \caption{\label{table:appendix:m4stat} Description / Statistics for the M4 competition dataset.}
    \vspace{5pt}
    \adjustbox{max width=\textwidth}{
        \begin{tabular}{l|rrrrrrrrr}
        \toprule
         \multicolumn{10}{c}{\hspace{1cm} Frequency / Horizon}\\
         \cmidrule{2-9}
         \multicolumn{2}{l}{Type}       & \multicolumn{2}{r}{Yearly / 6} & Quarterly / 8 & Monthly / 18 & Weekly / 13 & Daily / 14 & Hourly / 48 & Total \\
         \midrule
         Demographic && \multicolumn{2}{r}{1,088} & 1,858 & 5,728 &   24 &   10 &    0 & 8,708 \\
         Finance     && \multicolumn{2}{r}{6,519} & 5,305 & 10,987 &  164 & 1,559 &    0 & 24,534 \\
         Industry    && \multicolumn{2}{r}{3,716} & 4,637 & 10,017 &    6 &  422 &    0 & 18,798 \\
         Macro       && \multicolumn{2}{r}{3,903} & 5,315 & 10,016 &   41 &  127 &    0 & 19,402 \\
         Micro       && \multicolumn{2}{r}{6,538} & 6,020 & 10,975 &  112 & 1,476 &    0 & 25,121 \\
         Other       && \multicolumn{2}{r}{1,236} &  865 &  277 &   12 &  633 &  414 & 3,437 \\
         \midrule
         Total       && \multicolumn{2}{r}{23,000} & 24,000 & 48,000 & 359 & 4,227 & 414 & 100,000 \\
         \midrule 
         Min. Length && \multicolumn{2}{r}{19} & 24 & 60 & 93 & 107 & 748& \\
         Max. Length && \multicolumn{2}{r}{841} & 874 & 2812 & 2610 & 9933 & 1008 & \\
         Mean Length && \multicolumn{2}{r}{37.3} & 100.2 & 234.3 & 1035.0 & 2371.4 & 901.9 & \\
         SD Length   && \multicolumn{2}{r}{24.5} & 51.1 & 137.4 & 707.1 & 1756.6 & 127.9 & \\
         \% Smooth   && \multicolumn{2}{r}{82\%} & 89\% & 94\% & 84\% & 98\% & 83\% & \\
         \% Erratic  && \multicolumn{2}{r}{18\%}& 11\% & 6\% & 16\% & 2\% & 17\% & \\
         \bottomrule
        \end{tabular} 
    }
    \end{center}
\end{table}

\begin{figure}
    \begin{center}
    \includegraphics[width=\textwidth]{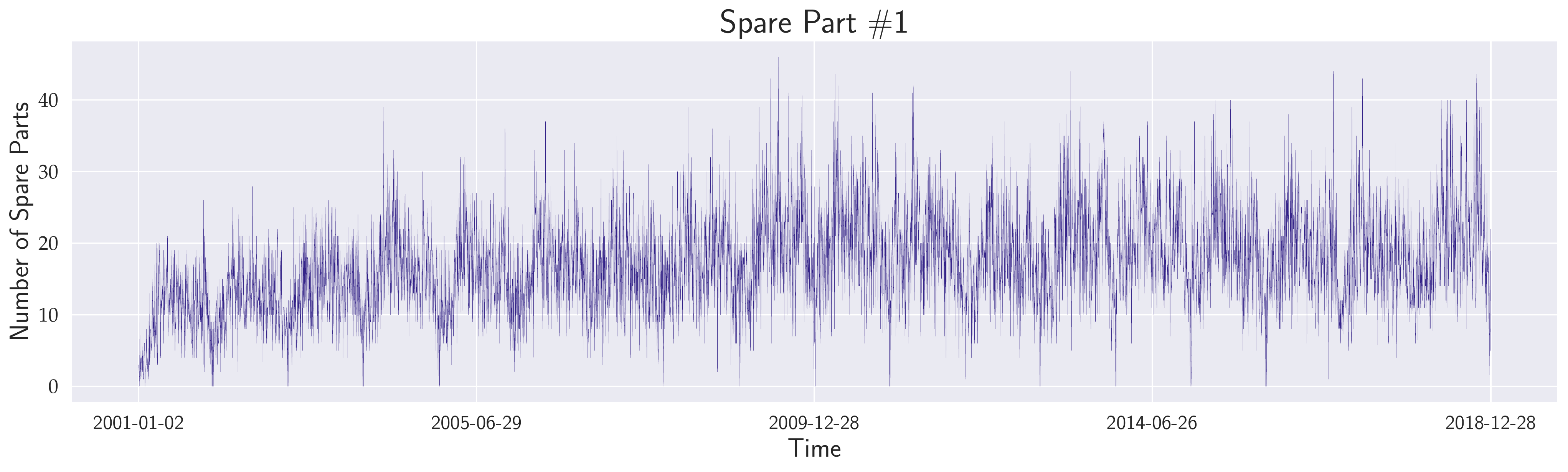}
    \includegraphics[width=\textwidth]{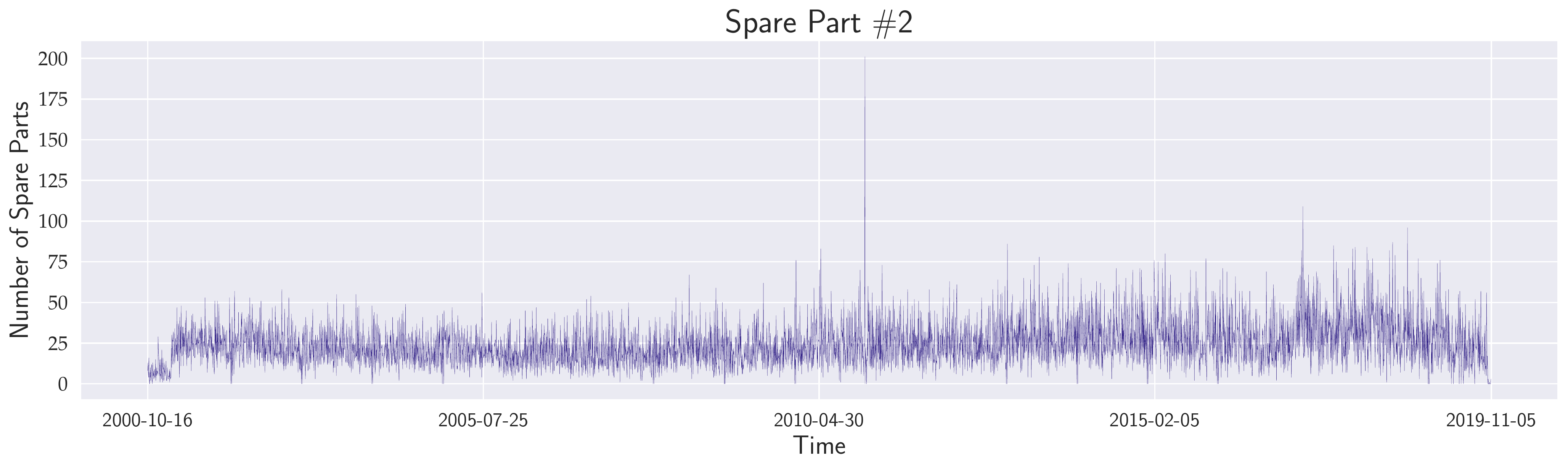}
    \includegraphics[width=\textwidth]{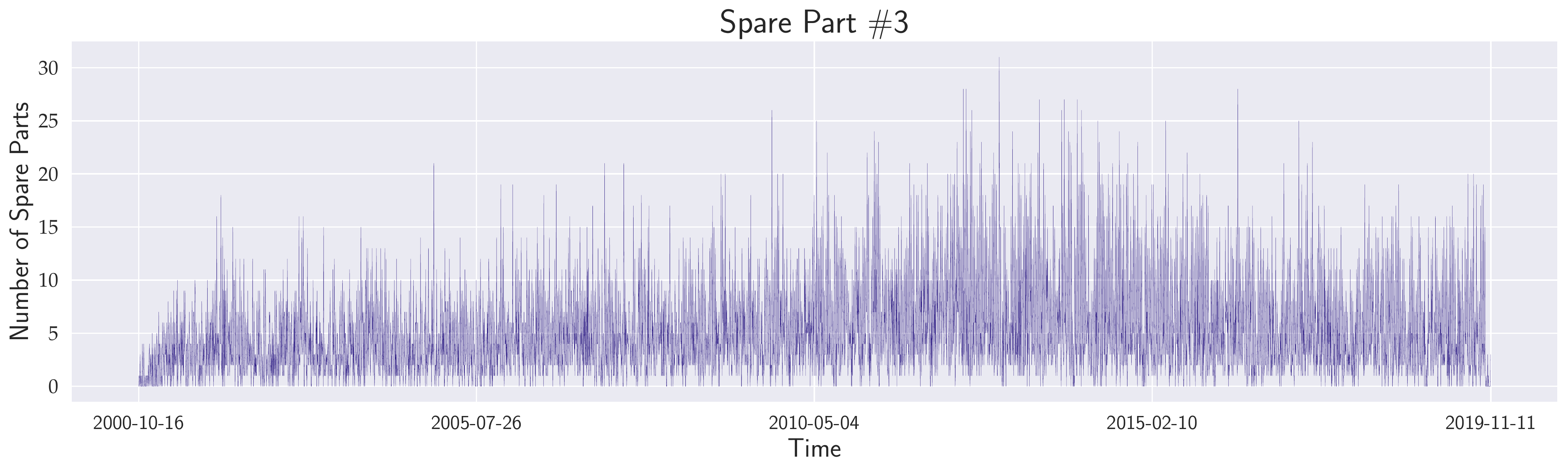}
    \includegraphics[width=\textwidth]{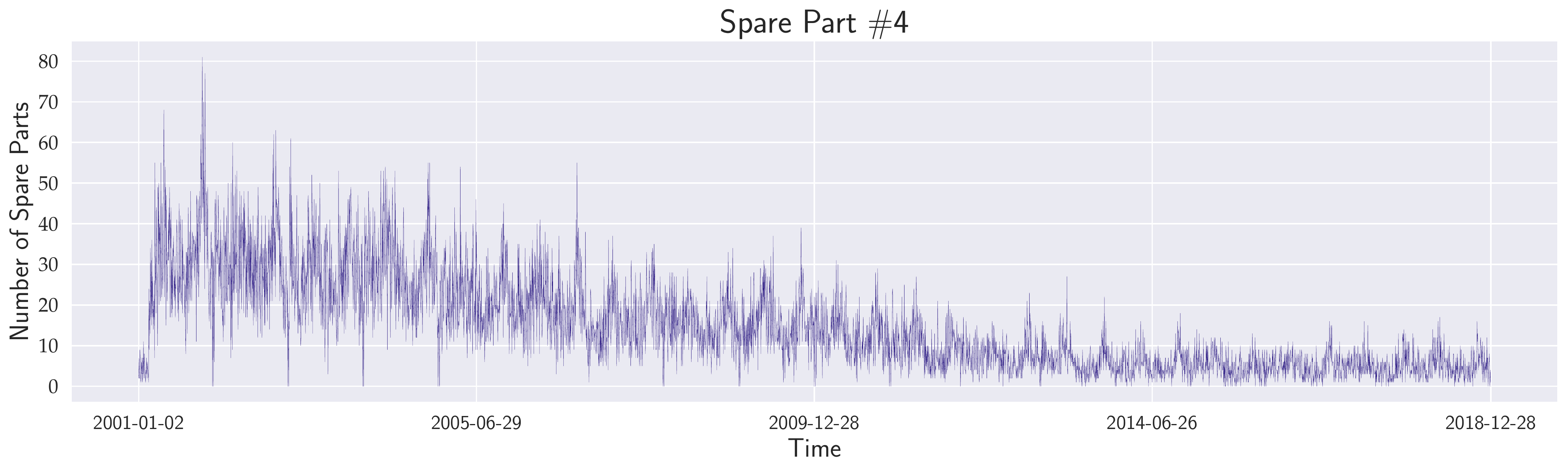}
    \caption{Visualization of the four proprietary \texttt{car-parts} time series.\label{fig:carparts}}
    \end{center}
\end{figure}

Table~\ref{table:appendix:single} lists key  statistics for the \texttt{car-parts} and the \texttt{electricity} time series we use in \textbf{[Manuscript, Section~\ref{subsection:single_time_series_experiments}]}. Notably, there are no time series categorized into the \emph{erratic} category, according to Syntetos et al. \cite{Syntetos05a}. As \texttt{car-parts} is proprietary, Fig.~\ref{fig:carparts} additionally shows a visualization of all observations from the four spare part demand time series.

\begin{table}[H]
    \begin{center}
        \caption{\label{table:appendix:single}Description / Statistics for the \texttt{car-parts} and \texttt{electricity} time series.}
        \vspace{5pt}
        \adjustbox{max width=\textwidth}{
            \begin{tabular}{lrr|cc}
                \toprule
                \multicolumn{3}{l}{}                    & \texttt{car-parts}                  & \texttt{electricity}  \\
                \midrule
                \multicolumn{2}{l}{Frequency / Horizon} &  & Daily / 1                        & 7Hourly / 1           \\
                \multicolumn{2}{l}{Total}               &  & 4                                & 10\footnotemark                  \\
                \multicolumn{2}{l}{Min. Length}         &  & 4507                             & 3762                  \\
                \multicolumn{2}{l}{Max. Length}         &  & 4783                             & 3762                  \\
                \multicolumn{2}{l}{Mean Length}         &  & 3644                             & 3762                  \\
                \multicolumn{2}{l}{SD Length}           &  & 137                               & 0                    \\
                \multicolumn{2}{l}{\% Smooth}           &  & 75\%                             & 70\%                  \\
                \multicolumn{2}{l}{\% Lumpy}            &  & 25\%                              & 30\%                 \\
                \bottomrule
            \end{tabular}
        }
    \end{center}
\end{table}
\footnotetext{In reference to \cite{Dua17a}, time series IDs are: MT$\_ i$ for $i \in \{14, 127, 130, 183, 238, 271, 318, 332, 333, 353\}.$   }
\newpage

\clearpage
\section{\textsc{Additional Results}}

Table~\ref{table:appendix:m4byapproaches} replicates \textbf{[Manuscript, Table~\ref{table:m4}]}, listing \texttt{sMAPE} / \texttt{OWA} statistics on M4, as well as the corresponding \texttt{MASE} / \texttt{OWA} statistics\footnote{Detailed results for \emph{Weekly}, \emph{Daily}  and \emph{Hourly} are not listed in \cite{Makridakis18b,Makridakis20a}, but available \href{https://github.com/Mcompetitions/M4-methods}{here}.}. Table~\ref{table:appendix:m4bydomain} lists results, split by time series \textbf{domains}. 
\begin{table}[H]
    \caption{\label{table:appendix:m4byapproaches}Performance comparison on the M4 benchmark in terms of (\subref{table:appendix:m4SMAPE}) 
    \textcolor{gred}{\texttt{sMAPE} / \texttt{OWA}} and (\subref{table:appendix:m4MASE}) \textcolor{gred}{\texttt{MASE} / \texttt{OWA}}, listed by
                subgroup. \texttt{N-BEATS} and \texttt{N-BEATS+TopAttn} denote an ensemble formed by training multiple
                models, varying $T$ from $2H$ to $5H$ and randomly initializing each model ten times (\ie, a total of 40
                models per subgroup). Forecasts are obtained by taking the median over the point forecasts of all
                models. $\dagger$ denotes results from \cite{Makridakis18b,Makridakis20a}. \label{table:m4_smape_owa}} 
    \begin{subtable}[t]{1.0\textwidth}
        \caption{\texttt{sMAPE} / \texttt{OWA}\label{table:appendix:m4SMAPE}}
        \vspace{-4pt}
    \begin{center}
            \adjustbox{max width=\textwidth}{
                \begin{tabular}{lrrrr|rr}
                \toprule
                \multirow{1}{*}{\textbf{Granularity}}        &
                \textbf{Total}                               &
                $^\dagger$\textbf{Winner M4}                 &
                $^\dagger$\textbf{Benchmark}                 &
                $^\dagger$\textbf{Naive2}                    &
                \textbf{\texttt{N-BEATS}} \cite{Oreshkin20a} &
                \textbf{\texttt{N-BEATS}}+\textcolor{gred}{\texttt{TopAttn}}                                                         \\
                \midrule
                Yearly                                       & (23k)                             & 13.176 / 0.778                & 14.848 / 0.867 & 16.342 / 1.000                & 13.149 / 0.776                                  & \textbf{13.063} / \textbf{0.771}                \\
                Quarterly                                    & (24k)                             & \hspace{0.09cm} 9.679 / 0.847 & 10.175 / 0.890 & 11.011 / 1.000                & \hspace{0.09cm} \textbf{9.684} / \textbf{0.845} & \hspace{0.09cm} 9.687 / \textbf{0.845}          \\
                Monthly                                      & (48k)                             & 12.126 / 0.836                & 13.434 / 0.920 & 14.427 / 1.000                & 12.054 / 0.829                                  & \textbf{12.025} / \textbf{0.828}                \\
                Weekly                                       & (359)                             & \hspace{0.09cm} 7.817 /  0.851 & \hspace{0.09cm}  8.944 / 0.926 & \hspace{0.09cm} 9.191 / 1.000 & \hspace{0.09cm} 6.447 / 0.703                   & \hspace{0.09cm} \textbf{6.361} / \textbf{0.699} \\
                Daily                                        & (4,227)                           & \hspace{0.09cm} 3.170 /  1.046 &  \hspace{0.09cm}  2.980 / 0.978            & \hspace{0.09cm} 3.045 / 1.000 & \hspace{0.09cm} \textbf{2.976} / \textbf{0.974} & \hspace{0.09cm} 2.979 / 0.975                   \\
                Hourly                                       & (414)                             & \hspace{0.09cm} \textbf{9.328} / \textbf{0.440}  & 22.053 / 1.556   & 18.383 / 1.000   & 10.040 / 0.464     & 10.271 /  0.483                                 \\
                \midrule
                \textbf{\textcolor{gblue}{Average}}          & \textcolor{gblue}{(100k)}         &
                \textcolor{gblue}{11.374 /  0.821}           & \textcolor{gblue}{12.555 / 0.898} &
                \textcolor{gblue}{13.564 / 1.000}            & \textcolor{gblue}{11.324 / 0.814} &
                \textbf{\textcolor{gblue}{11.291}} / \textbf{\textcolor{gblue}{0.811}}                                                                            \\
                \bottomrule
            \end{tabular}
            }
    \end{center}
    \vspace{-2pt}
\end{subtable}

\begin{subtable}[t]{1.0\textwidth}
    \begin{center}
        \caption{\texttt{MASE} / \texttt{OWA}\label{table:appendix:m4MASE}} 
        \adjustbox{max width=\textwidth}{
            \begin{tabular}{lrrrr|rr}
            \toprule
            \multirow{1}{*}{\textbf{Granularity}}        &
            \textbf{Total}                               &
            $^\dagger$\textbf{Winner M4}                 &
            $^\dagger$\textbf{Benchmark}                 &
            $^\dagger$\textbf{Naive2}                    &
            \textbf{\texttt{N-BEATS}} \cite{Oreshkin20a} &
            \textbf{\texttt{N-BEATS}}+\textcolor{gred}{\texttt{TopAttn}}                                                                                                                                                                                                          \\
            \midrule 
            Yearly                                       & (23k)                             & 2.980 / 0.778                & 3.280 / 0.867 & 3.974 / 1.000                & 2.972 / 0.776                                  & \textbf{2.950} / \textbf{0.771}                 \\
            Quarterly                                    & (24k)                             & 1.118 / 0.847                & 1.173 / 0.890 & 1.371 / 1.000                & \textbf{1.111} / \textbf{0.845}                & 1.112 / \textbf{0.845}                 \\
            Monthly                                      & (48k)                             & 0.884 / 0.836                &  0.966 / 0.920 & 1.063 / 1.000                & 0.875 / 0.829                                  & \textbf{0.874} / \textbf{0.828}                \\
            Weekly                                       & (359)                             & 2.356 / 0.851                &  2.432  /  0.926  & 2.777 / 1.000                & \textbf{1.950} / 0.703                         & 1.953 / \textbf{0.699}                          \\
            Daily                                        & (4,227)                           & 3.446 / 1.046                &  3.203 / 0.978  & 3.278 / 1.000                & \textbf{3.183} / \textbf{0.974}                & 3.188 / 0.975                   \\
            Hourly                                       & (414)                             & \textbf{0.893} / \textbf{0.440}  &  4.582 / 1.556  & 2.395 / 1.000               & 0.917 / 0.464                 & 0.974 /  0.483                                 \\
            \midrule
            \textbf{\textcolor{gblue}{Average}}          & \textcolor{gblue}{(100k)}         &
            \textcolor{gblue}{1.536 /  0.821}           & \textcolor{gblue}{1.663 / 0.898} &
            \textcolor{gblue}{1.912 / 1.000}            & \textcolor{gblue}{1.516 / 0.814} &
            \textbf{\textcolor{gblue}{1.511}} / \textbf{\textcolor{gblue}{0.811}}                                                                                                                                                                                                \\
            \bottomrule
        \end{tabular}
    }
\end{center}    
\end{subtable}
\end{table}

\vskip -0.5cm
\begin{table}[H]
    \begin{center}
            \caption{\label{table:appendix:m4bydomain}Performance comparison on the M4 benchmark in terms of (\subref{table:appendix:m4DOMAINSMAPE}) \texttt{sMAPE}$_{\text{\texttt{N-BEATS}}}$ / \texttt{sMAPE}$_{\text{\texttt{N-BEATS+TopAttn}}}$ 
            and (\subref{table:appendix:m4DOMAINMASE}) \texttt{MASE}$_{\text{\texttt{N-BEATS}}}$ / \texttt{MASE}$_{\text{\texttt{N-BEATS+TopAttn}}}$, listed by subgroup and \textbf{domain}. \texttt{N-BEATS} and \texttt{N-BEATS+TopAttn} denote the same models as in Table~\ref{table:appendix:m4byapproaches}.} 
        \begin{subtable}[t]{1.0\textwidth}
            \caption{\label{table:appendix:m4DOMAINSMAPE} 
            \texttt{sMAPE}$_{\text{\texttt{N-BEATS}}}$ / \texttt{sMAPE}$_{\text{\texttt{N-BEATS+TopAttn}}}$}
        \adjustbox{max width=\textwidth}{
                \begin{tabular}{lrrrrrr}
                \toprule
                \multirow{2}{*}{\textbf{Granularity}}        &
                Demographic                                  &
                Finance                                      &
                Industry                                     &
                Macro                                        &
                Micro                                        &
                Other                                                      \\
                & (8,7k) & (24,5k) & (18,8k) & (19,4k) & (25,1k) & (3,5k)  \\
                \midrule
                Yearly                                       &  9.640 / 9.694 & 14.029 / 13.879 & 16.645 / 16.523 & 13.450 / 13.400 & 10.700 / 10.654 & 13.094 / 13.000 \\
                Quarterly                                    &  9.908 / 9.933 & 11.158 / 11.161 & \hspace{0.09cm} 8.822 / \hspace{0.09cm} 8.832   & \hspace{0.09cm} 9.182 / \hspace{0.09cm} 9.178   & \hspace{0.09cm} 9.919 / \hspace{0.09cm} 9.922   & \hspace{0.09cm} 6.222 / \hspace{0.09cm} 6.173   \\
                Monthly                                      &  4.605 / 4.599 & 13.629 / 13.625 & 12.918 / 12.913 & 12.490 / 12.428 & 13.180 / 13.122 & 11.987 / 11.932 \\
                Weekly                                       &  1.401 / 1.403 & \hspace{0.09cm} 7.598 / \hspace{0.09cm} 7.516   & \hspace{0.09cm} 2.563 / \hspace{0.09cm} 2.548   & 11.303 / 10.837 & \hspace{0.09cm} 3.658 / \hspace{0.09cm} 3.681   & 12.204 / 12.112 \\
                Daily                                        &  6.300 / 6.313 &  \hspace{0.09cm} 3.442 / \hspace{0.09cm} 3.446  &  \hspace{0.09cm} 3.831 / \hspace{0.09cm} 3.832  & \hspace{0.09cm} 2.532 / \hspace{0.09cm} 2.532   & \hspace{0.09cm} 2.288 / \hspace{0.09cm} 2.291   & \hspace{0.09cm} 2.901 / \hspace{0.09cm} 2.901   \\
                Hourly                                       &                &                 &                 &                 &                 & \hspace{0.09cm} 9.787 / \hspace{0.09cm} 9.997   \\
                \midrule
                \textbf{\textcolor{gblue}{Average}}          &  \textbf{\textcolor{gblue}{6.358}} / \textcolor{gblue}{6.367} & \textcolor{gblue}{12.513} / \textbf{\textcolor{gblue}{12.472}} & \textcolor{gblue}{12.437} / \textbf{\textcolor{gblue}{12.413}} &
                                                                                                  \textcolor{gblue}{11.709} / \textbf{\textcolor{gblue}{11.665}} & \textcolor{gblue}{11.070} / \textbf{\textcolor{gblue}{11.034}} & \hspace{0.09cm} \textcolor{gblue}{8.997} / \hspace{0.09cm} \textbf{\textcolor{gblue}{8.971}}                                                                                                         \\
                \bottomrule
            \end{tabular}
        }
    \end{subtable}
    \begin{subtable}[t]{1.0\textwidth}
        \caption{\label{table:appendix:m4DOMAINMASE} 
        \texttt{MASE}$_{\text{\texttt{N-BEATS}}}$ / \texttt{MASE}$_{\text{\texttt{N-BEATS+TopAttn}}}$}
        \adjustbox{max width=\textwidth}{   
                \begin{tabular}{lrrrrrr}
                \toprule
                \multirow{2}{*}{\textbf{Granularity}}                       &
                Demographic                                                 &
                Finance                                                     &
                Industry                                                    &
                Macro                                                       &
                Micro                                                       &
                Other                                                       \\
                & (8,7k) & (24,5k) & (18,8k) & (19,4k) & (25,1k) & (3,5k)   \\
                \midrule
                Yearly                                                      & 2.410 / 2.428 & 3.086 / 3.055 & 3.021 / 2.996 & 2.956 / 2.932 & 2.994 / 2.981 & 2.647 / 2.616  \\
                Quarterly                                                   & 1.234 / 1.238 & 1.110 / 1.111 & 1.075 / 1.077 & 1.123 / 1.121 & 1.128 / 1.128 & 0.866 / 0.861 \\
                Monthly                                                     & 0.864 / 0.862 & 0.912 / 0.912 & 0.936 / 0.935 & 0.878 / 0.876 & 0.790 / 0.788 & 0.780 / 0.778  \\
                Weekly                                                      & 1.782 / 1.839 & 1.661 / 1.634 & 3.724 / 3.808 & 2.042 / 2.114 & 2.393 / 2.399 & 0.910 / 0.899  \\
                Daily                                                       & 9.604 / 9.641 & 3.396 / 3.402 & 3.784 / 3.787 & 3.198 / 3.205 & 2.597 / 2.603 & 3.519 / 3.520  \\
                Hourly                                                      &               &               &               &               &               & 0.903 / 0.971  \\
                \midrule
                \textbf{\textcolor{gblue}{Average}}                         & \textbf{\textcolor{gblue}{1.148}} / \textcolor{gblue}{1.151} & \textcolor{gblue}{1.695} / \textbf{\textcolor{gblue}{1.687}} & \textcolor{gblue}{1.447} / \textbf{\textcolor{gblue}{1.443}} &
                                                                                                          \textcolor{gblue}{1.380} / \textbf{\textcolor{gblue}{1.374}} & \textcolor{gblue}{1.558} / \textbf{\textcolor{gblue}{1.554}} & \textcolor{gblue}{1.993} / \textbf{\textcolor{gblue}{1.989}}                                                                                            \\
                \bottomrule
            \end{tabular}
        }
    \end{subtable}

    \end{center}
\end{table}

\section{\textsc{Hyperparameter Settings}}

Hyperparameter settings for our single time series experiments of \textbf{[Manuscript, Section~\ref{subsection:single_time_series_experiments}]} and the large-scale M4 experiments of \textbf{[Manuscript, Section~\ref{subsection:large_scale_experiments_on_the_M4_benchmark}]} are listed in Tables~\ref{table:appendix:singletsparams} and \ref{table:appendix:M4params}. 

\begin{table}[H]
    \begin{center}
       \caption{\emph{Single time series experiment} hyperparameters for \texttt{car-parts} and \texttt{electricity} data.\label{table:appendix:singletsparams}} \vskip4pt
        \adjustbox{max width=\textwidth}{
            \begin{tabular}{lrr|cc}
                \toprule
                \multicolumn{2}{l}{}           & & \texttt{car-parts}  & \texttt{electricity}    \\
                \multicolumn{2}{l}{Parameters} & & Daily               & 7Hourly                 \\
                \midrule
                \multicolumn{2}{l}{Iterations} & & 2k                  & 1.5k                    \\
                \multicolumn{2}{l}{Loss}       & & \multicolumn{2}{c}{\texttt{MSE}}                \\
                \multicolumn{2}{l}{$H$ (Forecast horizon)}   & & \multicolumn{2}{c}{1}          \\
                \multicolumn{2}{l}{Lookback period(s), $T$}   & & \multicolumn{2}{c}{14$H$ - 244$H$ }      \\
                \multicolumn{2}{l}{Batch size}             & & \multicolumn{2}{c}{30}                   \\
                \multicolumn{2}{l}{Attention heads}   & & \multicolumn{2}{c}{4}                    \\
                \multicolumn{2}{l}{Barcode coordinate functions} & & \multicolumn{2}{c}{32}                   \\
                \multicolumn{2}{l}{Encoder-layers}    & & \multicolumn{2}{c}{1}                    \\
                \multicolumn{2}{l}{Hidden dimension}  & & \multicolumn{2}{c}{128}                  \\
                \bottomrule
            \end{tabular}
        }
    \end{center}
\end{table}

As mentioned in the manuscript, for M4 experiments with \texttt{N-BEATS} (and \texttt{N-BEATS+TopAttn}), we closely follow the \emph{generic} \texttt{N-BEATS} parameter configuration of Oreshkin et al. \cite[Table 18]{Oreshkin20a}; any additional parameters (for our \texttt{N-BEATS+TopAttn} approach) are highlighted in \textcolor{gred}{red}.
Note that we also mark \emph{Hidden dimension} in \textcolor{gred}{red}, as this is not only the hidden dimension of the \texttt{N-BEATS} blocks, but we equally use this setting for the hidden dimension of the transformer encoder layers.

\begin{table}[H]
    \begin{center}
        \caption{\emph{Large-scale experiment} hyperparameters across all subsets of the \texttt{M4} dataset. Parameters specific to \texttt{N-BEATS+TopAttn} are highlighted in \textcolor{gred}{red}. For a detailed description of the \texttt{N-BEATS} parameters, we refer to \cite[Section~D.1]{Oreshkin20a}.
        \label{table:appendix:M4params}} \vskip4pt
    \adjustbox{max width=\textwidth}{
        \begin{tabular}{lrr|rrrrrrr}
            \toprule
            \multicolumn{2}{c}{} && \multicolumn{6}{c}{M4} \\
            \multicolumn{2}{l}{Parameters}&& Yearly & Quarterly & Monthly & Weekly & Daily & Hourly \\
            \midrule
            \multicolumn{2}{l}{$H$ (Forecast horizon)} && 6 & 8 & 18 & 13 & 14 & 48                  \\
            \multicolumn{2}{l}{$L_{H}$} && 1.5 & 1.5 & 1.5 & 10 & 10 & 10                  \\
            \multicolumn{2}{l}{Iterations} && \multicolumn{6}{c}{5k}                       \\
            \multicolumn{2}{l}{Loss}       && \multicolumn{6}{c}{\texttt{sMAPE}}           \\
            \multicolumn{2}{l}{Lookback period(s), $T$} && \multicolumn{6}{c}{2$H$, 3$H$, 4$H$, 5$H$}      \\
            \multicolumn{2}{l}{Batch size} && \multicolumn{6}{c}{1024}                          \\
            \multicolumn{2}{l}{\textcolor{gred}{Attention heads}} && \multicolumn{6}{c}{2}                   \\
            \multicolumn{2}{l}{\textcolor{gred}{Barcode coordinate functions}} && \multicolumn{6}{c}{64}                \\
            \multicolumn{2}{l}{\textcolor{gred}{Encoder-layers}} && \multicolumn{6}{c}{20}                   \\
            \multicolumn{2}{l}{\textcolor{gred}{Hidden dimension}}&& \multicolumn{6}{c}{128}                \\
            \multicolumn{2}{l}{Double-Residual Blocks} && \multicolumn{6}{c}{1}            \\
            \multicolumn{2}{l}{Block-layers} && \multicolumn{6}{c}{4}                      \\
            \multicolumn{2}{l}{Stacks} && \multicolumn{6}{c}{30}                           \\
            \bottomrule
        \end{tabular}
    }
    \end{center}
\end{table}

\clearpage

\section{\textsc{Sliding Window Configurations}}
Lets assume we have, at one point in training, a randomly extracted training portion of $T+H$ consecutive observations from a length-$N$ time series 
($T \ll N$). We use the first $T$ observations as (1) \emph{raw} input signal $\boldsymbol{x}$ to our models and (2) for extraction of complementary \emph{local topological properties}. The $H$ consecutive observations, starting at $T+1$, are used as target (to compute the mean-squared-error, or the \texttt{sMAPE} for instance). 

\emph{Throughout all experiments, sliding windows are moved forward by one observation a time.}

For extracting local topological properties from $\boldsymbol{x}$ (of length $T$) via persistent homology, two parameters are necessary: the parameter $W$ determines the number of overlapping sliding windows and the parameter $n$ determines the length of a single sliding window (i.e., $n$ observations). For each sliding window, we obtain one barcode (or two, if $-\boldsymbol{x}$ is taken into account). 

\textbf{Singe time series experiments [Manuscript, Section~\ref{subsection:single_time_series_experiments}].} In this setting, $H=1$, as we compute one-step forecasts. Since, typically, forecast models differ in their sensitivity to the length $T$ of the
input observations $\boldsymbol{x}$, we cross-validate $T$ (for all methods) using the \texttt{sMAPE} on the validation set. 

The collection of $T$ used for cross-validation is constructed based on the following consideration: first, for persistent homology computation, we need a reasonable amount of observations in each sliding window; and, second, we need a reasonable amount of sliding windows for self-attention. Hence, we choose (1) $W \geq 10$ and (2) $n\leq45$. For one specific choice of $(W,n)$, we get $T = W + n - 1$. Varying $W \in \{5, 25, 45\}$ and $n \in \{10, 20, 50, 70, 100, 150, 200, 232\}$ thus determines the length, $T$, of the input vector $\boldsymbol{x}$. For instance, setting $(W,n) = (5,10)$ gives a decomposition of $\boldsymbol{x}$ (of length 14), into $5$ subsequent windows of length $10$ for which persistent homology is computed. Overall, in the described setup, $T$ ranges from 14 to 244.

\textbf{Large-scale experiments [Manuscript, Section~\ref{subsection:large_scale_experiments_on_the_M4_benchmark}].} In this setting, $H>1$. For comparability with \texttt{N-BEATS}, we stick to the original setup of considering input lengths as multiples of the forecast horizon (which is specific to each subgroup in M4). In particular, $T$ ranges from $2H$ to $5H$, see Table~\ref{table:appendix:M4params}. As an example, on M4 \emph{Yearly}, this yields a range of $T$ from 12 to 30. As mentioned in \textbf{[Manuscript, Section~\ref{subsection:forecasting_model_large_scale}]}, we set $n = \lfloor 0.7 \cdot T \rfloor$ and $W$ is thus determined by $(T,n)$.
\clearpage

\section{\textsc{Ensemble Size}}

As described in \textbf{[Manuscript, Section~\ref{subsection:forecasting_model_large_scale}]}, we ensemble 40 models to obtain forecasts for each subgroup of the M4 dataset. One ensemble is formed per subgroup and consists of training \texttt{N-BEATS}, or \texttt{N-BEATS+TopAttn}, respectively, with 10 random initializations for four different values of $T$, i.e., $2H,3H,4H,5H$ (where $H$ denotes the specific forecast horizon prescribed per subgroup), using the \texttt{sMAPE} as a loss function. In case of \emph{Yearly} for instance, $H=6$, see Table~\ref{table:appendix:m4stat}. 

\begin{figure}[h!]
\begin{center}
    \includegraphics[scale=0.4]{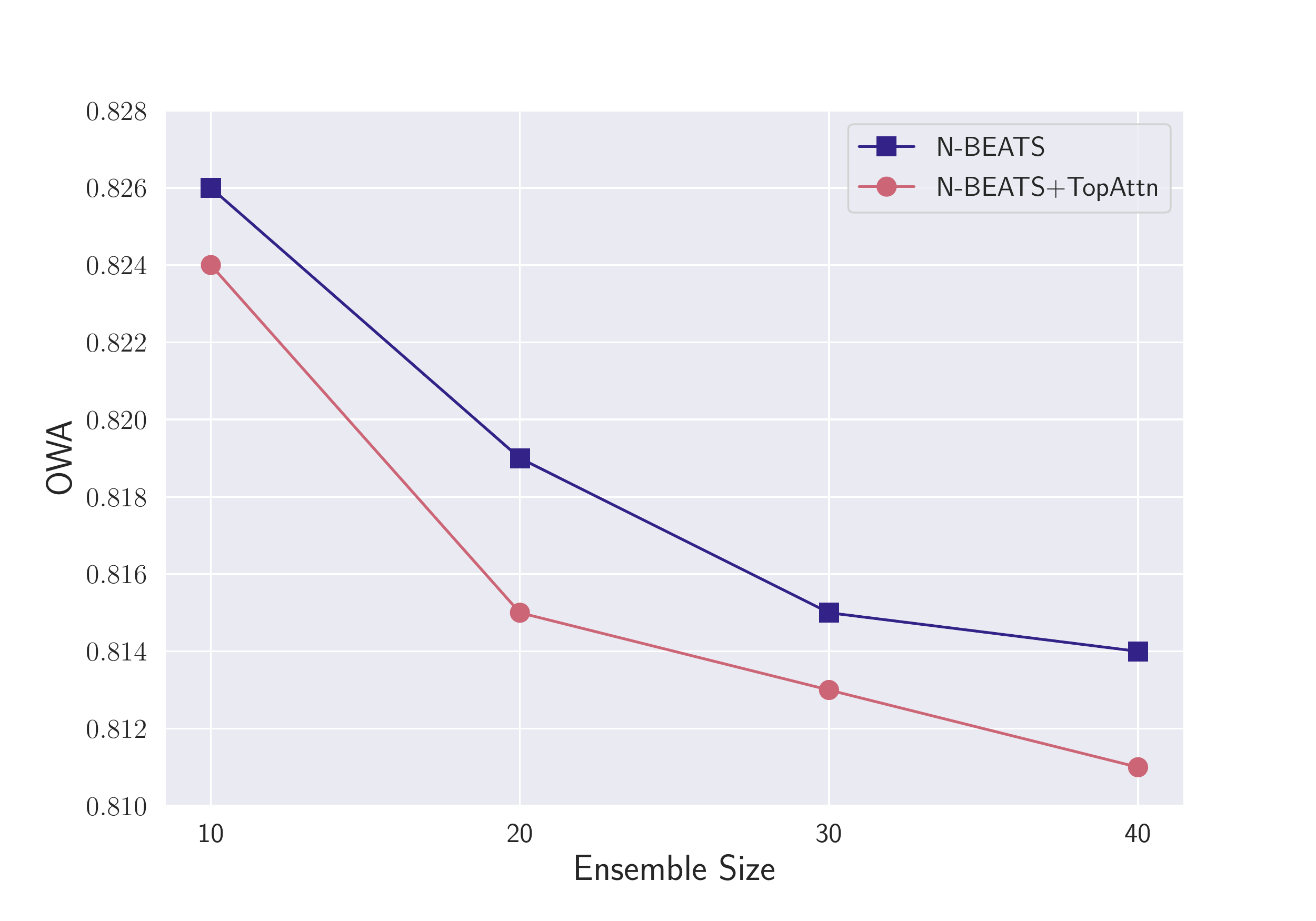}
\end{center}
\caption{\label{fig:appendix:ensemblecomp}Comparison of \texttt{N-BEATS} and  \texttt{N-BEATS+TopAttn} in terms of varying the ensemble size. At the maximum ensemble size of 40, the \texttt{OWA} corresponds to the \texttt{OWA} reported in, e.g., Table~\ref{table:appendix:m4byapproaches}.}
\end{figure}

Fig.~\ref{fig:appendix:ensemblecomp} shows a comparison of \texttt{N-BEATS} and  \texttt{N-BEATS+TopAttn} over the ensemble size, illustrating the \texttt{N-BEATS+TopAttn} equally benefits from a larger ensemble. Notably, in \cite{Oreshkin20a} the ensemble is much larger, as, in addition to training models with the \texttt{sMAPE} as loss, the \texttt{MAPE} and \texttt{MASE} are used and $T$ scales up to $7H$, resulting in 180 models in total. 

\section{\textsc{Model Details}}

In this section, we describe the details for the models used in the
\textit{single time series} experiments of \textbf{[Manuscript, Section~\ref{subsection:single_time_series_experiments}]}. 

\textbf{Prophet.} We use the publicly available Python implementation of \texttt{Prophet}\footnote{\url{https://facebook.github.io/prophet/}} with \emph{default} parameter choices.

\textbf{autoARIMA.}
We use the publicly available Python implementation of \texttt{autoARIMA}\footnote{\url{https://alkaline-ml.com/pmdarima/}}. In terms of hyperparameters, the initial number of time lags of the auto-regressive (``AR'') and the moving-average
(``MA'') model is set to 1 bounded by its maximum 6. The period for seasonal differencing is equal to 5; the order of first-differencing and of seasonal differencing is set to 2 and 0, respectively.

\textbf{LSTM.}
We implement a \texttt{LSTM} model with hidden dimensionality 128, 8 recurrent layers and a dropout layer on the outputs of each \texttt{LSTM} layer with dropout probability of 0.3. Outputs of the LSTM are fed to a subsequent single-hidden-layer \texttt{MLP} with hidden dimensionality equal to 64, including batch normalization and ReLU activation. Initial learning rate and weight decay are set to 1e-3 and 1.25e-5, respectively. We minimize the mean-squared-error (MSE) via \textsc{ADAM} over 1.5k (\texttt{electricity}) and 2k
(\texttt{car-parts}) iterations, respectively, using a batch size of 30. All learning rates are annealed following a cosine learning rate schedule.

For \textbf{DeepAR}, \textbf{MQ-CNN}, \textbf{MQ-RNN} and the \textbf{MLP} baseline, we use the publicly available \texttt{GluonTS} \cite{gluonts_jmlr} implementations\footnote{\url{https://ts.gluon.ai}}, mostly with \emph{default} parameter choices. We only adjust the number of (maximum) training epochs to 20 (for comparability to our approach, where we count iterations), change the hidden dimensionality of the \textbf{MLP} to 64 and set the batch size to 30.

\section{\textsc{System Setup}}
\label{appendix:section:system}

All experiments were executed on an Ubuntu Linux 20.04 system, using \texttt{PyTorch} v1.7.0 (CUDA 10.1), 128 GB of RAM and 16 Intel(R) Core(TM) i9-10980XE CPUs.

\section{\textsc{Persistent Homology Runtime}}

\begin{figure}[h!]
\centering
\includegraphics[scale=0.4]{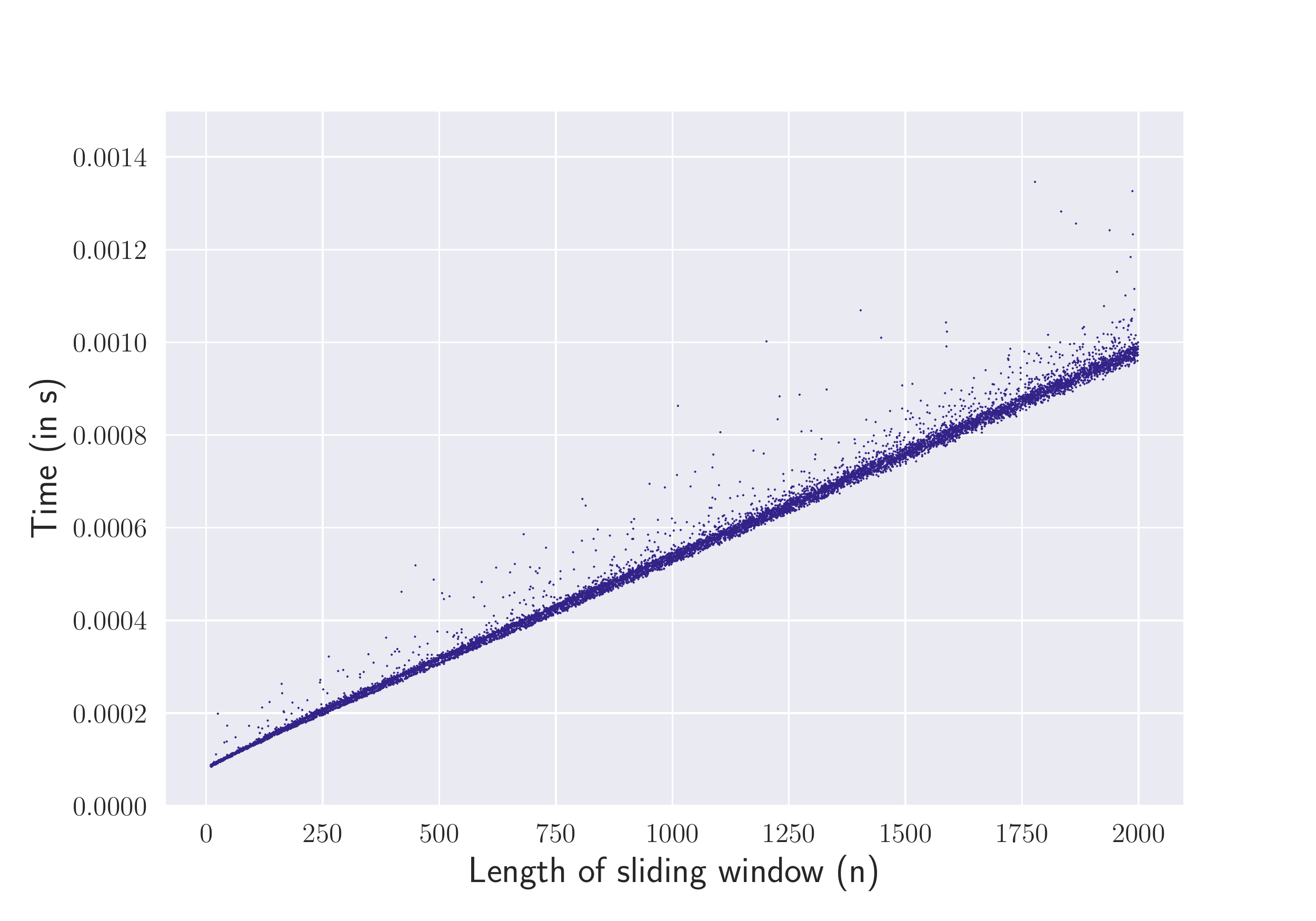}
\caption{\label{fig:appendix:runtime}Runtime (in seconds) for 0-dimensional persistent homology computation from observations within a sliding window, varying in length ($n$) from $[5,2000]$.}
\end{figure}

To back up the ``near-linear runtime'' statement for 0-dimensional persistent homology computation in the proposed regime (see \textbf{[Manuscript, Section~\ref{subsection:forecasting_model}]}), Fig.~\ref{fig:appendix:runtime} shows a runtime plot (using \texttt{Ripser}\footnote{\url{https://github.com/Ripser/ripser}}) over 10,000 sliding window sizes, $n$, in the range $[5,2000]$. The system setup for these runtime experiments is given in Section~\ref{appendix:section:system}. Fig.~\ref{fig:appendix:runtime} clearly corroborates the statement from the manuscript.

